\documentclass[lettersize,journal]{IEEEtran}
\usepackage{amsmath,amsfonts}
\usepackage{algorithmic}
\usepackage{algorithm}
\usepackage{array}
\usepackage[caption=false,font=normalsize,labelfont=sf,textfont=sf]{subfig}
\usepackage{textcomp}
\usepackage{stfloats}
\usepackage{url}
\usepackage{verbatim}
\usepackage{graphicx}
\usepackage{cite}
\usepackage{multirow}
\usepackage{amssymb}
\usepackage{multicol}
\usepackage{booktabs}
\usepackage{xcolor}
\usepackage{colortbl}
\usepackage{listings}
\usepackage{pifont}
\usepackage{bbding}
\usepackage{color}
\usepackage[colorlinks,
            linkcolor=red,
            anchorcolor=blue,
            citecolor=green
            ]{hyperref}

\begin{document}

\title{SCASeg: Strip Cross-Attention for Efficient Semantic Segmentation}

\author{
Guoan Xu, 
Jiaming Chen,
Wenfeng Huang,
Wenjing Jia,~\IEEEmembership{Member,~IEEE}, 
Guangwei Gao,~\IEEEmembership{Senior Member,~IEEE}, 
and Guo-Jun Qi,~\IEEEmembership{Fellow,~IEEE}
\thanks{This work was supported in part by the Foundation of the State Key Laboratory of Integrated Services Networks of Xidian University under Grant No. ISN27-4.~\textit{ (Corresponding authors: Wenjing Jia, Guangwei Gao.)}}
\thanks{Guoan Xu, Wenfeng Huang, and Wenjing Jia are with the Faculty of Engineering and Information Technology, University of Technology Sydney, Sydney, NSW 2007, Australia, and Guoan Xu is also with the State Key Laboratory of Integrated Services Networks, Xidian University, Xi’an 710071, China (e-mail: xga\_njupt@163.com, huang-wenfeng@outlook.com, and Wenjing.Jia@uts.edu.au).}
\thanks{Jiaming Chen is with the Department of Computer Science, The University of Manchester, Manchester, Oxford Rd, Manchester M13 9PL, United Kingdom (e-mail: ppjmchen@gmail.com).}
\thanks{Guangwei Gao is with the PCA Lab, Key Laboratory of Intelligent Perception and Systems for High-Dimensional Information of Ministry of Education, School of Computer Science and Engineering, Nanjing University of Science and Technology, Nanjing 210094, China, and also with the State Key Laboratory of Integrated Services Networks, Xidian University, Xi’an 710071, China (e-mail: gwgao@njust.edu.cn).}
\thanks{Guo-Jun Qi is with the Research Center for Industries of the Future and the School of Engineering, Westlake University, Hangzhou 310024, China, and also with OPPO Research, Seattle, WA 98101 USA (e-mail: guojunq@gmail.com).}
}

\markboth{IEEE Transactions on Image Processing}%
{Shell \MakeLowercase{\textit{et al.}}: A Sample Article Using IEEEtran.cls for IEEE Journals}


\maketitle

\begin{abstract}
The Vision Transformer (ViT) has achieved notable success in computer vision, with its variants widely validated across various downstream tasks, including semantic segmentation. However, as general-purpose visual encoders, ViT backbones often do not fully address the specific requirements of task decoders, highlighting opportunities for designing decoders optimized for efficient semantic segmentation. This paper proposes Strip Cross-Attention (SCASeg), an innovative decoder head specifically designed for semantic segmentation. Instead of relying on the conventional skip connections, we utilize lateral connections between encoder and decoder stages, leveraging encoder features as Queries in cross-attention modules. Additionally, we introduce a Cross-Layer Block (CLB) that integrates hierarchical feature maps from various encoder and decoder stages to form a unified representation for Keys and Values. The CLB also incorporates the local perceptual strengths of convolution, enabling SCASeg to capture both global and local context dependencies across multiple layers, thus enhancing feature interaction at different scales and improving overall efficiency. To further optimize computational efficiency, SCASeg compresses the channels of queries and keys into one dimension, creating strip-like patterns that reduce memory usage and increase inference speed compared to traditional vanilla cross-attention. Experiments show that SCASeg's adaptable decoder delivers competitive performance across various setups, outperforming leading segmentation architectures on benchmark datasets, including ADE20K, Cityscapes, COCO-Stuff 164k, and Pascal VOC2012, even under diverse computational constraints.
\end{abstract}


\begin{figure}[htbp]
	\centerline{\includegraphics[width=9cm]{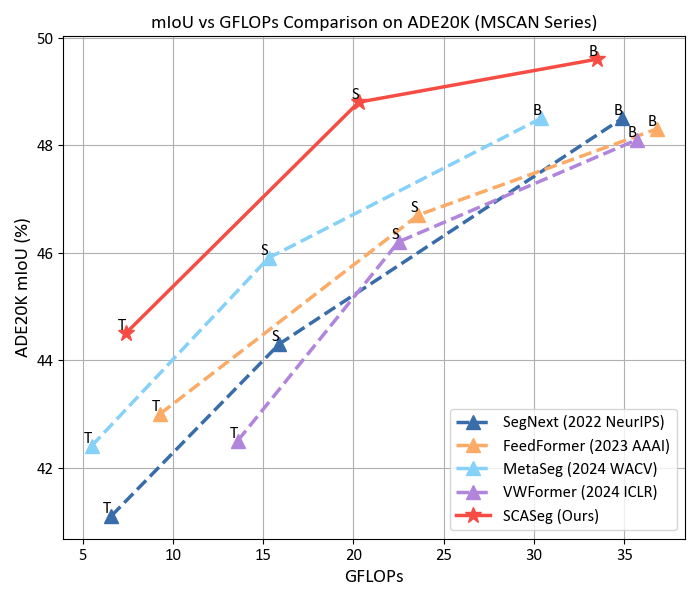}}
	\caption{The mIoU and GFLOPs comparison of SCASeg with SOTA approaches on ADE20K~\cite{zhou2017scene} dataset. The results are reported using a single model and single-scale inference based on MSCAN~\cite{guo2022segnext} backbones.}
	\label{vs}
\end{figure}

\section{Introduction}

\IEEEPARstart{S}{emantic} segmentation is a fundamental task in computer vision that involves pixel-level classification~\cite{yuan2020object,shi2022transformer,ma2024efficient}. This process entails labeling each pixel in an image to accurately identify object categories, spatial positions, and other critical information, thereby providing a detailed understanding of the scene's composition. Semantic segmentation has widespread applications in various fields, including autonomous driving~\cite{xu2023lightweight}, medical diagnosis~\cite{azad2024medical}, and remote sensing~\cite{wang2024samrs}, among others~\cite{kirillov2023segment,lai2024lisa}. A pivotal development in this area was the introduction of the fully convolutional network (FCN)~\cite{long2015fully},  which popularized the encoder-decoder architecture. In this architecture, the encoder extracts high-level semantic features while the decoder integrates these features with spatial details. Despite subsequent advancements~\cite{fu2019dual}, traditional CNNs still struggle to capture long-range dependencies effectively.

This limitation has been largely addressed with the emergence of Transformers~\cite{vaswani2017attention}. Following its groundbreaking success in Natural Language Processing (NLP), the Transformer quickly began to make a significant impact on vision tasks as well. Its self-attention mechanism is highly effective at capturing long-range dependencies within input sequences. Dosovitskiy \textit{et al.}~\cite{dosovitskiy2020image} extended this concept to the visual domain by proposing the Vision Transformer (ViT) backbone, which involved dividing images into small patches, transforming them into one-dimensional sequences, and feeding these sequences into the Transformer encoder to align the input dimensions for visual processing. Since then, many researchers have integrated ViT into the semantic segmentation domain, yielding impressive results. 
However, most of these approaches have focused primarily on optimizing the efficiency of the Transformer encoder while relying on simple or pre-existing designs for the decoder architecture. For instance, SegFormer~\cite{xie2021segformer} emphasizes designing an efficient Transformer encoder while utilizing a straightforward all-MLP decoder. Similarly, SegNeXt~\cite{guo2022segnext} develops a lightweight and efficient backbone but adopts a simplistic approach in the decoder stage. 

More recently, MetaSeg~\cite{kang2024metaseg} introduced a new and efficient self-attention module called Channel Reduction Attention (CRA), which simplifies the channel dimensions of the query and key into a single dimension per head within the self-attention process. 
However, this approach does not effectively facilitate interaction among the various feature representations. FeedFormer~\cite{shim2023feedformer} uses features directly as queries, rather than relying on class-specific learnable queries. 
U-MixFormer~\cite{yeom2025u} adaptively incorporates multi-stage features as keys and values within its specialized mix-attention module. 
MacFormer~\cite{xu2024macformer} introduces a mutual agent cross-attention mechanism to enhance bidirectional feature interaction. Additionally, it proposes detailed enhancements in the frequency domain, achieving notable results. 
Despite the advancements offered by these attention blocks, they do not adequately consider the importance of local information. As demonstrated by models such as Metaformer~\cite{yu2022metaformer}, CMT~\cite{guo2022cmt}, SMT~\cite{lin2023scale}, and XCiT~\cite{ali2021xcit}, convolution methods are more effective than Transformers in capturing local features. Therefore, it is crucial to integrate local perception capabilities into the model alongside global attention mechanisms.

Based on the above observations 
and aiming to achieve a balance between efficiency and performance, we propose a novel Cross-Layer Block (CLB) comprising a \textit{Strip Cross-Attention (SCA)} module and a \textit{Local Perception Module (LPM)}. Specifically, the SCA module captures 
global long-range context dependencies by employing 
a low-rank strategy to compress the channel dimensions of the query and key for lightweight computation, while retaining 
the full value dimension to maintain dense token connectivity. This approach effectively balances global attention modeling 
with manageable computational complexity. Meanwhile, the LPM leverages the local perception capability of convolution, enhanced 
with channel attention, to extract and retain 
fine-grained local details. This design 
achieves a better trade-off between efficiency and effectiveness, as illustrated in Fig.~\ref{vs}. 
In summary, the main contributions of our method are as follows:
\begin{enumerate}

\item 
We present an innovative and robust transformer-decoder architecture designed for efficient semantic segmentation. Building on U-Net's strengths in capturing and transmitting hierarchical features, our approach uniquely utilizes lateral connections from the transformer encoder as query features. 

\item 
We introduce a meticulously designed Cross-Layer Block consisting of two key modules: Strip Cross-Attention and the Local Perception Module. These modules work together to capture both local and global contexts effectively.
 
\item 
Experiments were conducted using various backbones on benchmark datasets, including ADE20K~\cite{zhou2017scene}, Cityscapes~\cite{cordts2016cityscapes}, COCO-Stuff 164k~\cite{caesar2018coco}, and Pascal VOC2012~\cite{hoiem2009pascal}, resulting in SOTA performance.
\end{enumerate}

\section{Related Work}

\subsection{Semantic Segmentation}

Semantic segmentation can be seen as an evolution of image classification, transitioning from categorizing entire images to assigning labels at the pixel level~\cite{liu2019auto,zheng2021rethinking,cheng2022masked,xu2024sctnet,xia2024vit}. During the deep learning era, the Fully Convolutional Network (FCN)~\cite{long2015fully} marked a significant breakthrough in semantic segmentation by utilizing a fully convolutional architecture for end-to-end pixel-wise classification. 
Following the development of FCN, 
subsequent research advanced semantic segmentation from several perspectives: 
1) Enlarging the receptive field~\cite{chen2018encoder}. DeepLab-v3~\cite{chen2017rethinking} introduced dilation rates in the Atrous Spatial Pyramid Pooling (ASPP) module, allowing for a larger and more multi-scale receptive field. 
2) Improving contextual information~\cite{zhang2018context}. CPNet~\cite{yu2020context} enhanced feature learning accuracy by encoding ground truth into a one-hot representation and introducing a context before the encoder, providing more precise guidance for feature learning. 
3) Incorporating boundary information~\cite{zhen2020joint}. BPKD~\cite{liu2024bpkd} employed edge detection operators to dilate and erode target objects, effectively extracting their edges, and used knowledge distillation to transfer accurate edge information from a teacher model to a student network. 
4) Designing various attention mechanisms~\cite{zhong2020squeeze}. DANet~\cite{fu2019dual} and CCNet~\cite{huang2019ccnet} extended non-local attention by 
integrating channel attention concepts to enhance overall model performance. 
Although these approaches have significantly improved segmentation accuracy, 
they have also introduced numerous empirical modules, resulting in computationally intensive and complex frameworks.

Another research direction explores prototype-driven designs (\textit{e.g.}, PEM~\cite{cavagnero2024pem}), which constrain cross-attention using learned prototypes and combine them with efficient multi-scale pyramids to reduce computational load. 
ContrastiveSeg~\cite{wang2021exploring} introduced pixel-wise metric learning to leverage semantics across images without additional overhead. 
Clustering-based approaches (\textit{e.g.}, CLUSTSEG~\cite{liang2023clustseg}) treated queries as cluster centers, alternating between pixel assignment and center updates to provide a unified framework for segmentation tasks. While these methods mainly innovate 
at the level of objectives or query formation, our work focuses on 
an architecture-level contribution through the design of a decoder and token mixer. 

\begin{figure*}[htbp]
	\centerline{\includegraphics[width=18cm]{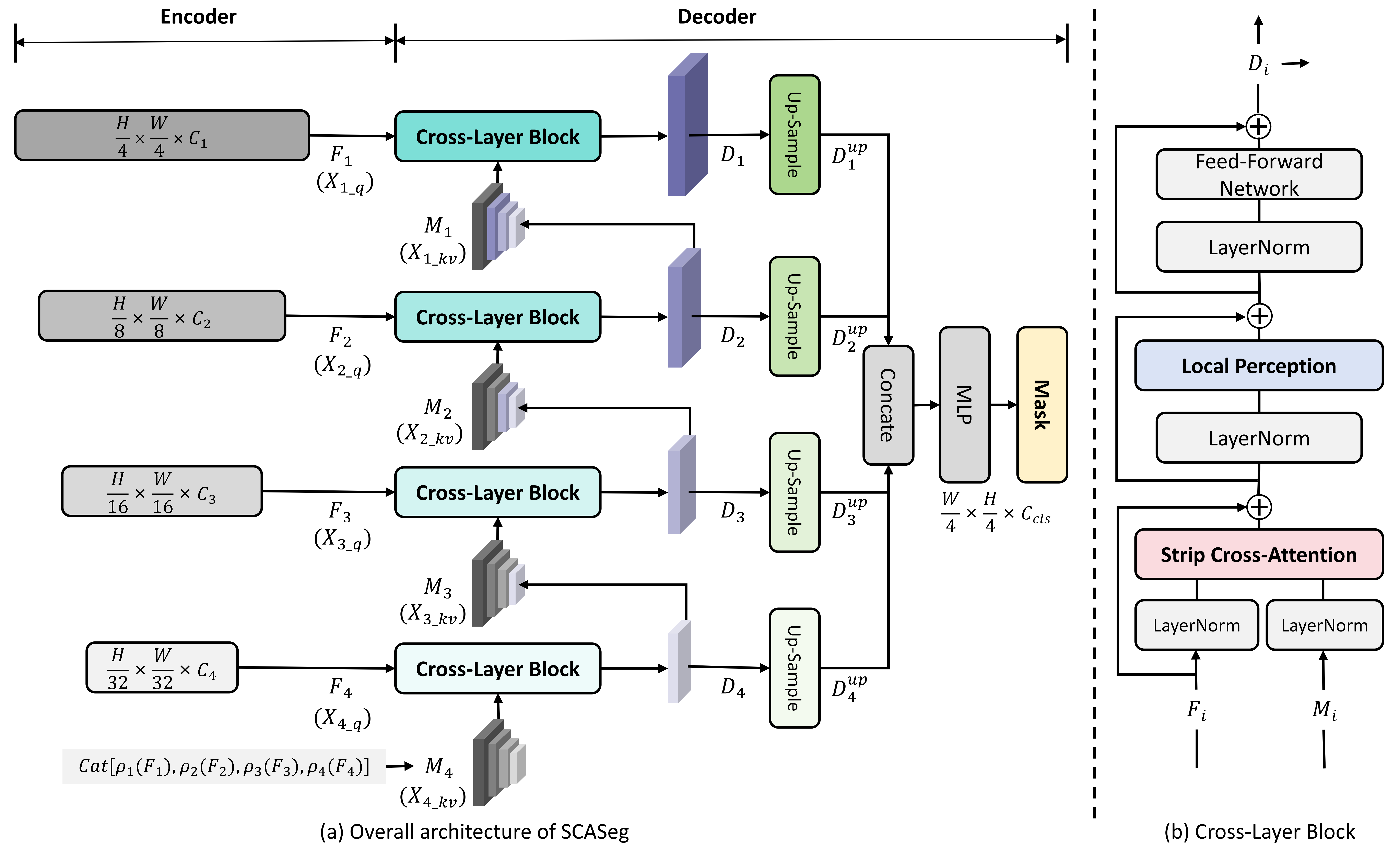}}
    \caption{
    The Overview of the proposed SCASeg architecture (a) and the detailed structure of the Cross-Layer Block (CLB) (b). SCASeg consists of two main components: an encoder based on hierarchical backbones (e.g., MSCAN~\cite{guo2022segnext} and MiT-Bx~\cite{xie2021segformer}) and a decoder with a cross-layer interaction design. The key innovation of SCASeg lies in this decoder structure, which promotes multi-scale feature complementarity. The \textbf{CLB} further enables efficient global–local feature integration while preserving local structural information.}
	\label{model}
    \vspace{-3mm}
\end{figure*}

\subsection{Encoder Backbone}

The Vision Transformer (ViT)~\cite{dosovitskiy2020image} was the pioneering work that demonstrated how a pure Transformer-based model could achieve SOTA performance in image classification. ViT treats each image as a sequence of tokens, which are processed through multiple Transformer layers for classification. Following this, DeiT~\cite{touvron2021training} introduced a more effective training strategy along with a distillation technique for ViT. Recent approaches~\cite{guo2022cmt,lin2023scale,ali2021xcit} have incorporated specialized modifications to ViT to further enhance its performance in image classification. 

However, since semantic segmentation involves dense prediction tasks distinct from image classification, improvements in classification models do not always translate to equivalent gains in segmentation performance. In this context, SETR~\cite{zheng2021rethinking} was the first to adopt a pure vision transformer-based architecture for semantic segmentation. However, its training costs are quite high due to the large number of tokens generated by high-resolution images, which substantially increases the computational complexity of the self-attention mechanism. Researchers have increasingly focused on improving the computational efficiency of the backbone. PVT~\cite{wang2021pyramid} employed spatial reduction in the linear projection stage to reduce the dimensions of the key and value matrices. The Swin Transformer~\cite{liu2021swin}, on the other hand, partitions images into multiple windows before applying patch sizes, significantly reducing the computational cost. Later, SegFormer~\cite{xie2021segformer} proposed a simple yet highly efficient structure that combined a Mix Transformer encoder with an all-MLP decoder. 
Subsequently, models like CoaT~\cite{xu2021co}, LeViT~\cite{graham2021levit}, and Twins~\cite{chu2021twins} further improved the continuity of local features and eliminated fixed-size positional embeddings to enhance Transformer performance in dense prediction tasks. In light of the high computational cost, SegNeXt~\cite{guo2022segnext} showed that convolutional attention offers a more efficient and effective means of encoding contextual information compared to the self-attention mechanism used in Transformers, while also providing a comprehensive analysis of the strengths and advantages of various models. 


\subsection{Decoder Head}


For semantic segmentation, Segmenter~\cite{strudel2021segmenter} leveraged the output embeddings associated with image patches and derived 
class labels from these embeddings using either a point-wise linear decoder or a mask transformer decoder. MetaSeg~\cite{kang2024metaseg} introduced a lightweight decoder module called Channel Reduction Attention, which enabled self-attention within each stage's output while reducing computational load. However, a key limitation is the lack of cross-layer interaction, indicating potential areas for improvement. FeedFormer~\cite{shim2023feedformer} enhanced efficiency by taking high-level encoder features as queries, and the lowest-level encoder features as keys and values. 
Yet, it processes feature maps independently without progressive propagation across decoder stages, missing opportunities for gradual refinement that can enhance object boundary detection. 
U-MixFormer~\cite{yeom2025u} addressed this by introducing a mix-attention mechanism that first downsampled features from different levels and concatenated them to form queries. Features from each encoder level were then treated as keys and values, with cross-attention applied progressively across layers. The newly generated feature map was merged back into the original concatenated features to form new queries. 

Our approach draws inspiration from this, but U-MixFormer~\cite{yeom2025u} still has a high computational cost due to its reliance on standard cross-attention mechanisms. On the other hand, MacFormer~\cite{xu2024macformer} preserved boundary information in the frequency domain and employs Mutual Agent cross-attention. While the use of agents helps control the overall parameter count, the bidirectional cross-attention operations still contribute to considerable computational complexity.

\begin{figure*}[htbp]
	\centerline{\includegraphics[width=18cm]{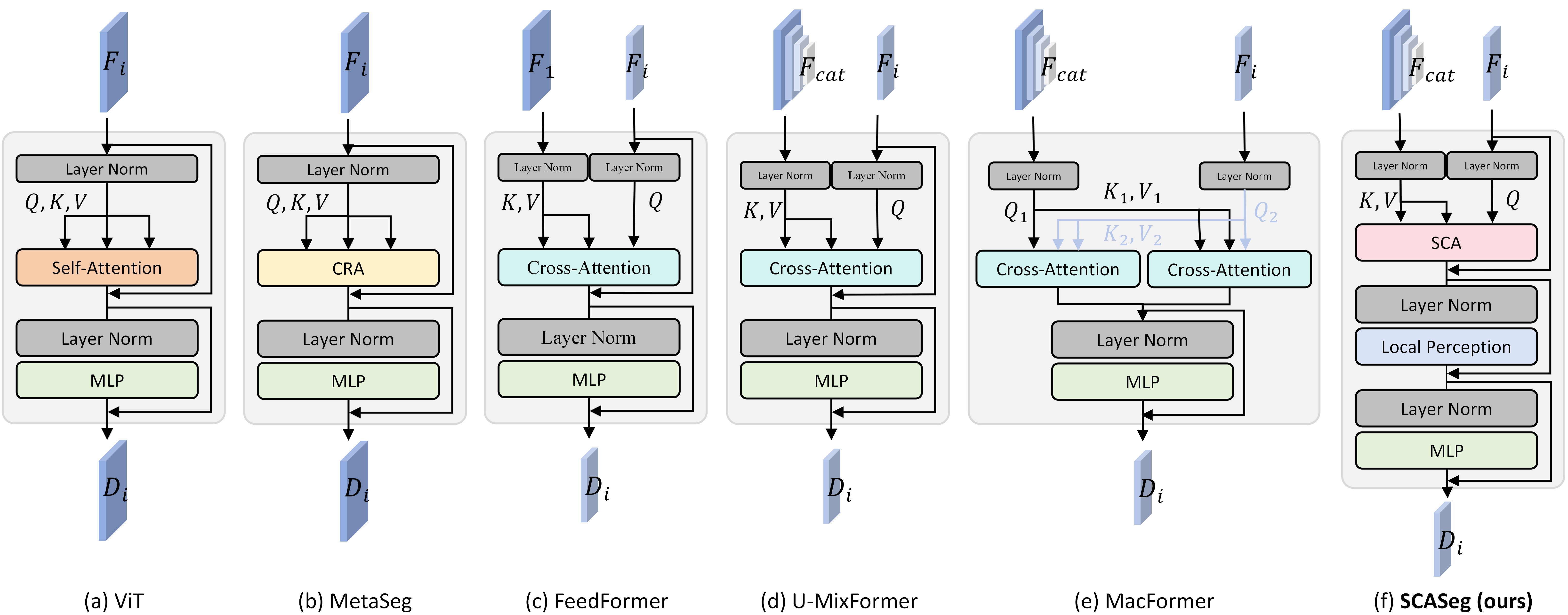}}
	\caption{The Cross-Layer Block (CLB) in our proposed SCASeg in comparison with its counterparts in SOTA approaches: (a) ViT~\cite{dosovitskiy2020image}, (b) MetaSeg~\cite{kang2024metaseg}, (c) FeedFormer~\cite{shim2023feedformer}, (d) U-MixFormer~\cite{yeom2025u}, (e) MacFormer~\cite{xu2024macformer}. The main distinction lies in the input features and the token mixer component. Additionally, our CLB takes into account the preservation of local information. }
	\label{attn}
    \vspace{-3mm}
\end{figure*}

\section{Methodology}

\subsection{Overall Architecture}
\label{sec31}

As illustrated in Fig.~\ref{model} (a), our SCASeg framework is compatible with any pretrained model that features a hierarchical four-stage architecture. In this work, we employ lightweight backbones such as MiT-B0$\sim$B5 and MSCAN-T/S/B as encoders, efficiently extracting rich feature representations. Inspired by U-MixFormer~\cite{yeom2025u}, we have developed a refined version of the U-Net structure for the decoder. This enhanced design integrates a Cross-Layer Block to facilitate inter-level feature interaction with a low computational burden, significantly improving decoding capability.

\subsection{Hierarchical Encoder and Lightweight Decoder}
\label{sec32}

Given an image $I$ of size $H \times W \times 3$ as input, feature maps $F_i \in \mathbb{R}^{\frac{H}{2^{i+1}} \times \frac{W}{2^{i+1}} \times C_i}$ are extracted at each stage of the encoder, where $i \in \{1, 2, 3, 4\}$ indicates the corresponding encoder stage and $C_i$ denotes the number of channels in that stage. These features provide a progression from coarse to fine detail, contributing to the improved performance of semantic segmentation.


The decoder in our SCASeg utilizes the U-Net architecture to better capture global contexts that are insufficiently addressed by the encoder. At each stage of the decoder, refined features $D_i$ are progressively generated through the Cross-Layer Block, where the query features $X^i_q$ correspond to the respective lateral encoder feature maps $F_i$. The key and value feature $X^i_{kv}$ (denoted as ${M_i}$ in Fig.~\ref{model}) are derived from a combination of both encoder and decoder stages. The decoder features are then upsampled using bilinear interpolation to match the height and width of $D_1$. Finally, the concatenated features are passed through an MLP (Multilayer Perceptron) to generate the segmentation masks, which have dimensions of $\frac{H}{4} \times \frac{W}{4} \times 3$. 

The entire decoding process can be summarized by the following formulas:
\begin{equation}
    \mathrm{M_i} = \mathrm{Cat}\left[ {{\mathrm{\rho _1}}\left( {{\mathrm{F_1}}} \right),...,{\rho _i}\left( {{\mathrm{F_i}}} \right),{\mathrm{D_{i + 1}}},...,\mathrm{{D_4}}} \right]_{i = 1}^4,
\end{equation}
\begin{equation}
    \mathrm{{D_i}} = \mathrm{CLB}\left( {\mathrm{{F_i}},\mathrm{{M_i}}} \right),
\end{equation}
\begin{equation}
    \mathrm{{D_i}}^\mathrm{{up}} = \mathrm{Up}\left( {\mathrm{{D_i}},{2^{i - 1}}} \right),
\end{equation}
\begin{equation}
\mathrm{{O_{mask}}} = \mathrm{MLP}(\mathrm{Cat}\left[ {\mathrm{D_i^{up}}} \right]_{i = 1}^4),
\end{equation}
where $Cat$ denotes the concatenation operation, $\rho$ represents a downsampling pooling operation, $CLB$ stands for Cross-Layer Block, and $Up$ refers to an upsampling function, which includes the scaling factor. The MLP is implemented using linear functions.

\begin{figure*}[ht]
	\centerline{\includegraphics[width=18cm]{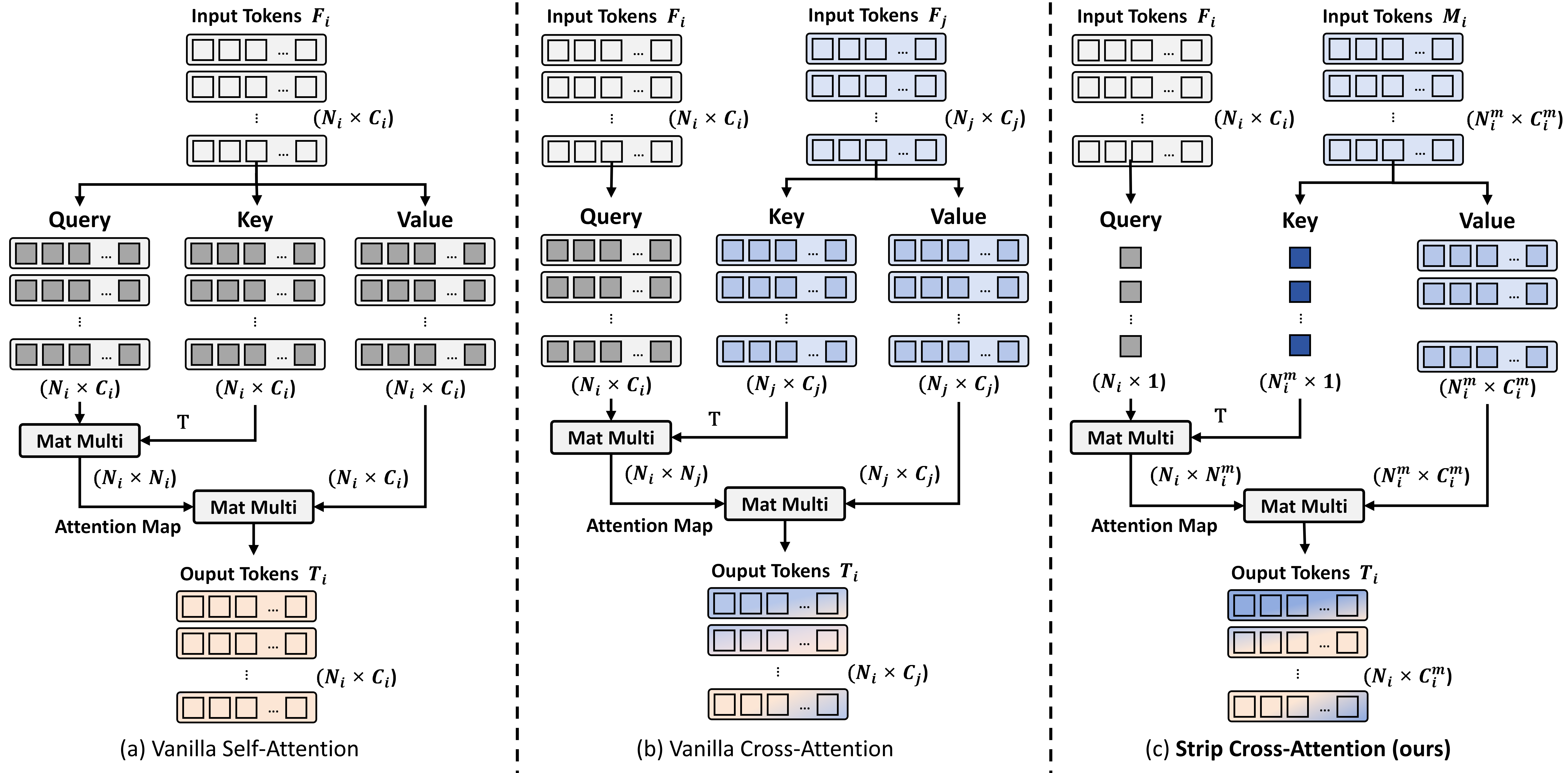}}
	\caption{The proposed Strip Cross-Attention (SCA) in comparison with the vanilla Self-Attention~\cite{dosovitskiy2020image} and Cross-Attention~\cite{chen2021crossvit}. The difference lies in the construction of the attention map: self-attention generates the attention map from the same feature, while cross-attention constructs it from different features. \textbf{Strip Cross-Attention (SCA)} leverages the advantages of cross attention while also reducing computational burden. Specifically, strip tokens are obtained by projecting query and key features into a single-channel embedding via linear transformation, resulting in a strip-shaped representation along the channel dimension. Unlike spatial aggregation methods, this operation preserves the spatial token resolution and keeps the token count unchanged, while reducing the dimensionality of attention similarity computation.} 
	\label{sca}
    \vspace{-3mm}
\end{figure*}

\subsection{Cross-Layer Block (CLB)}
\label{sec34}

The proposed Cross-Layer Block (CLB) incorporates the MetaFormer~\cite{yu2022metaformer} block in the decoder to enhance the global context of the feature representations extracted by the encoder, with a primary focus on integrating contextual information across different hierarchical features. As illustrated in Fig.~\ref{model} (b), the CLB includes the MetaFormer block, which consists of three residual subblocks, a Local Perception Module (LPM), and a novel Strip Cross-Attention (SCA) module for token mixing. The SCA module effectively captures both local and global contexts of the features while seamlessly integrating information across different hierarchical levels with minimal computational cost. The CLB is applied at each stage and takes two distinct inputs, $F_i$ and $M_i$. 

Thus, the entire process is defined as follows:
\begin{equation}
    \mathrm{Z_i^G} = \mathrm{SCA}\left( {\mathrm{LN}\left( {\mathrm{{F_i}}} \right),\mathrm{LN}\left( {\mathrm{{M_i}}} \right)} \right) + \mathrm{{F_i}},
\end{equation}
\begin{equation}
    \mathrm{Z_i^{GL}} = \mathrm{LPM}\left( {\mathrm{LN}\left( \mathrm{{Z_i^G}} \right)} \right) + \mathrm{Z_i^G},
\end{equation}
\begin{equation}
    \mathrm{{D_i}} = \mathrm{MLP}\left( {\mathrm{LN}\left( \mathrm{{Z_i^{GL}}} \right)} \right) + \mathrm{Z_i^{GL}},
\end{equation}
where ${Z_i}^G$ captures global features, whereas ${Z_i}^{GL}$ fuses both local and global contexts. Layer Normalization ($LN$) is employed to standardize these features.

The relationship to other attention blocks is shown in Fig.~\ref{attn}. These works introduce various modifications to the self-attention block,  with a focus on enhancing the token-mixer component. MetaSeg~\cite{kang2024metaseg} addresses the computational complexity of self-attention by employing a channel reduction strategy for $Q$ and $K$. However, it overlooks the complementary information across multi-scale features. On the other hand, U-MixFormer~\cite{yeom2025u} utilizes a feature mixing strategy but fails to capture local information. MacFormer~\cite{xu2024macformer} incorporates bidirectional complementarity but imposes a heavy computational burden. Considering these factors, we aim to propose a more efficient approach that balances both computational efficiency and feature complementarity. These methods often prioritize one aspect over others, resulting in limitations regarding feature interactions, computational cost, or local context preservation. In contrast, our method effectively addresses these challenges by integrating the advantages of cross-layer fusion, local context retention, and computational efficiency.



\subsection{Strip Cross-Attention (SCA)}
\label{sec35}

We introduce the Strip Cross-Attention (SCA) module as an innovative token mixer within the CLB, which is designed to effectively handle both global and local feature extraction while maintaining computational efficiency in cross-attention for semantic segmentation tasks. In transformer blocks, attention modules calculate the scaled dot-product attention for queries ($Q$), keys ($K$), and values ($V$) using the following formula:
\begin{equation}
\mathrm{Attn}(\mathbf{Q},\mathbf{K},\mathbf{V})=\mathrm{Softmax}(\frac{\mathbf{Q}\mathbf{K}^\top}{\sqrt{\mathrm{d_k}}})\mathbf{V}.
\end{equation}
Here, $\sqrt {{\mathrm{d_k}}}$ represents the dimension of the key embeddings.

\begin{table*}[!t]
\caption{Performance comparisons with SOTA light-weight models on ADE20K~\cite{zhou2017scene} and Cityscapes~\cite{cordts2016cityscapes}. Results are taken from the original papers under a unified mmsegmentation training protocol. “–” indicates metrics not reported in the original publications.}
\begin{center}	
\scalebox{1.0}{
\begin{tabular}{l||lcc|cccc}
\toprule
\multirow{2}{*}{\textbf{Method}} & \multirow{2}{*}{\textbf{Year}} & \multirow{2}{*}{\textbf{Backbone}} & \multirow{2}{*}{\textbf{Params. (M)}$\downarrow$} & \multicolumn{2}{c}{\textbf{ADE20K}~\cite{zhou2017scene}} & \multicolumn{2}{c}{\textbf{Cityscapes}~\cite{cordts2016cityscapes}} \\
&   &   &   & \textbf{GFLOPs}$\downarrow$ & \textbf{mIoU (\%)}$\uparrow$ & \textbf{GFLOPs}$\downarrow$ & \textbf{mIoU (\%)}$\uparrow$ \\ \midrule

FCN~\cite{long2015fully}   & 2015 CVPR  & MobileNet-V2  & 9.8  & 39.6  & 19.7  
& 317.1  & 61.5      \\
PSPNet~\cite{zhao2017pyramid}  & 2017 CVPR & MobileNet-V2 & 13.7  & 52.9  & 29.6   
& 423.4  & 70.2        \\
DeepLabV3+~\cite{chen2018encoder}  & 2018 ECCV & MobileNet-V2  & 15.4  & 69.4  
& 34.0   & 555.4  & 75.2    \\
SwiftNetRN~\cite{cheng2019swiftnet}  & 2019 arxiv   & ResNet-18   & 11.8   & -   & - & 104.0  & 75.5   \\
Semantic FPN~\cite{li2023convmlp}   & 2021 CVPR& ConvMLP-S & 12.8  & 33.8   
& 35.8   & -  & -   \\

\midrule
LeMoRe~\cite{abid2025lemore} & 2025 arxiv & LeMoRe & \textbf{1.6} & \underline{0.8} & 33.5 & - & -\\
DataFormer~\cite{abid2025dataformer} & 2025 CVPR & DataFormer & \textbf{1.6} & \textbf{0.6} & 33.8 & - & -\\
ContextFormer~\cite{abid2025contextformer} & 2025 arxiv & TPEM & \underline{1.7} & \textbf{0.6} & 35.0 & - & -\\
SeaFormer~\cite{wanseaformer}
& 2023 ICLR & SeaFormer-S 
& 4.0 & 1.1 & 38.1 & - & 76.1  \\
SCTNet~\cite{xu2024sctnet} 
& 2024 AAAI & SCTNet-S & 4.7 & - & 37.7 & - & 72.8  \\
SDPT~\cite{cao2024sdpt} & 2024 TITS & SDPT-Tiny 
& 3.6 & 5.7 & 39.4 
&\textbf{63.4} & 77.3   \\

SegFormer~\cite{xie2021segformer}  & 2021 NeurIPS & MiT-B0 & 3.8 & 8.4  & 37.4   
& 125.5  & 76.2  \\
FeedFormer~\cite{shim2023feedformer}   & 2023 AAAI  & MiT-B0  & 4.5  & 7.8  & 39.2  
& 107.4  & 77.9   \\
PEM~\cite{cavagnero2024pem} & 2024 CVPR & STDC1 & 17.0 & 16.0 & 39.6 & 92.0 & \underline{78.3} \\
MetaSeg~\cite{kang2024metaseg}    & 2024 WACV   & MiT-B0  & 4.1   & 3.9   & 37.9  & \underline{90.9}  & 76.7  \\
VWFormer~\cite{yan2024multiscale} & 2024 ICLR & MiT-B0 & 3.7 & 5.8 & 38.9 & 172.0 & 77.2  \\
EMOv2-2M~\cite{zhang2025emov2} & 2025 TPAMI & MiT-B0 &2.6 &10.3 &40.2  & -& - \\
 
\midrule
\rowcolor{gray!20}SCASeg (Ours) & \textbf{-}   & MiT-B0   & 6.0  & 5.9   & \textbf{41.6}   & 101.7  & \textbf{79.3}   \\ 
\midrule

SegFormer~\cite{xie2021segformer}    & 2021 NeurIPS  & LVT & \textbf{3.9}  & 10.6  & 39.3   
& 140.9 & 77.6    \\
FeedFormer~\cite{shim2023feedformer}   & 2023 AAAI & LVT & \underline{4.6}  & 10.0  & 41.0 
& 124.6  & 78.6    \\
PEM~\cite{cavagnero2024pem} & 2024 CVPR & STDC-1 & 17.0 & 16.0 
& 39.6 & \textbf{92.0} & 78.3 \\
MetaSeg~\cite{kang2024metaseg} & 2024 WACV   & LVT  & 4.2 & \underline{6.0}   &  40.8 & \underline{106.0}  & 78.1  \\
CCASeg~\cite{yoo2025ccaseg} & 2025 WACV &MiT-B0 &6.2 &7.2 &42.6 &115.8 &78.7  \\
EDAFormer~\cite{yu2024embedding} & 2024 ECCV & EDAFormer-T & 4.9 & \textbf{5.6} & \underline{42.3} & 151.7 & 78.7\\ 
VWFormer~\cite{yan2024multiscale} & 2024 ICLR & LVT & 5.3 & 14.3 & \underline{42.3} & 194.0 & \underline{78.9}  \\ \midrule
\rowcolor{gray!20}SCASeg (Ours)  & \textbf{-}   & LVT   & 6.3   & 8.8 & \textbf{43.8}  & 122.4  & \textbf{79.7}   \\ 
\midrule
SegNeXt~\cite{guo2022segnext}     & 2022 NeurIPS   & MSCAN-T  & \textbf{4.3}   & \underline{6.6}   & 41.1  & 56.0  & 79.8    
\\
FeedFormer~\cite{shim2023feedformer}   & 2023 AAAI & MSCAN-T & \underline{5.0}  & 9.3  & 43.0 
& 61.1  & \underline{80.6}    \\
MetaSeg~\cite{kang2024metaseg}    & 2024 WACV   & MSCAN-T  & 4.7   & \textbf{5.5}   & 42.4  & \textbf{47.9}  & 80.1    
\\
PEM~\cite{cavagnero2024pem} & 2024 CVPR & STDC2 & 21.0 & 19.3 & \textbf{45.0} & 118.0 & 79.0 \\
VWFormer~\cite{yan2024multiscale} & 2024 ICLR & MSCAN-T & 5.8 & 13.6 & 42.5 & 131.4 & 80.3  \\
VRWKV~\cite{duanvision} & 2025 ICLR & VRWKV-T & 8.4 &16.6 &43.3 &- &-\\

\midrule
\rowcolor{gray!20}SCASeg (Ours)  & \textbf{-}   & MSCAN-T   & 6.5  & 7.4  & \underline{44.5}  & \underline{54.8}  & \textbf{81.2}  \\ 
\bottomrule
\end{tabular}
}
\label{light}
\vspace{-3mm}
\end{center}
\end{table*}

\begin{figure}[t]
	\centerline{\includegraphics[width=0.9\linewidth]
    {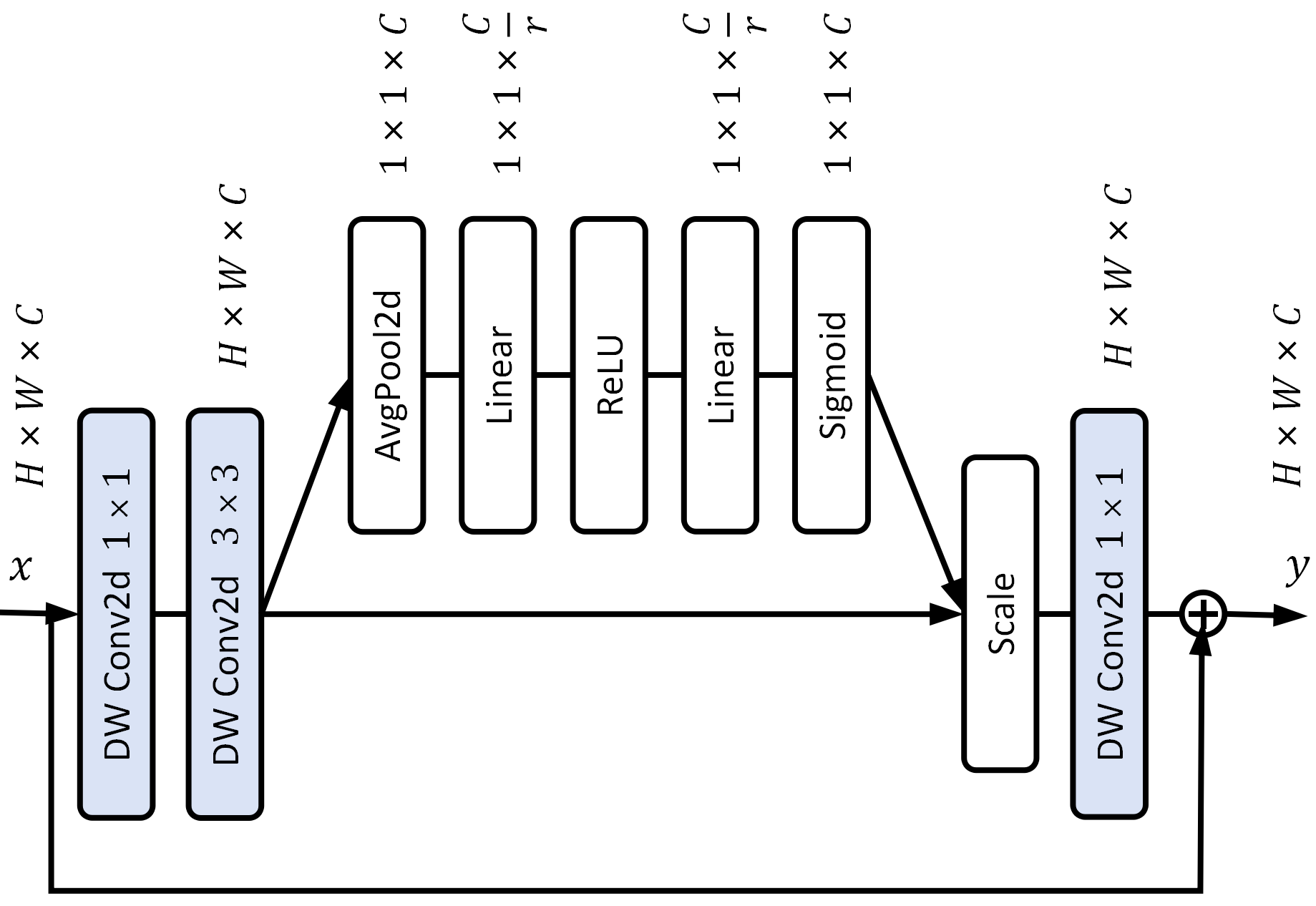}}
	\caption{The architecture of our proposed Local Perception Module (LPM).}
	\label{lpm}
    \vspace{-3mm}
\end{figure}

In Self-Attention, the features used to generate the queries, keys, and values are identical (denoted as $X_{qkv}$) and are derived from a common input source $F_i$, as shown in Fig.~\ref{sca} (a). In Cross-Attention, two distinct sets of features ($X_{q}$ and $X_{kv}$) are processed, each originating from a separate source, $F_i$ and $F_j$, as depicted in Fig.~\ref{sca} (b). In Strip Cross-Attention, a fused feature ($X_{kv}$) is gathered from multiple stages at different scales, denoted as $M_i$. This design enables the query to identify matches across various stages with varying levels of contextual granularity, thus supporting improved feature refinement. From an efficiency perspective, we strategically designed strip-shaped tokens to implement the attention map. The channel dimensions of the original query and key are embedded into a single dimension, further reducing computational overhead from $\mathcal{O}(H\cdot N^2\cdot C)$ (vanilla cross-attention) to $\mathcal{O}(H \cdot N^2\cdot 1)$. 
This one-dimensional transformation significantly reduces computational complexity. The comparison formulas for the computational costs of Self-Attention and Strip Cross-Attention are as follows:
\begin{equation}
    \Omega(\mathrm{SA})=\mathrm{N^2} \cdot \mathrm{C}+\mathrm{N^2} \cdot \mathrm{C}
\end{equation}
\begin{equation}
    \Omega(\mathrm{SCA})=\mathrm{N^2} \cdot \mathbf{1} +\mathrm{N^2} \cdot \mathrm{C},
\end{equation}
where $N$ represents the total number of tokens.

Motivated by MetaSeg~\cite{kang2024metaseg}, we observed that the channel-compressed feature token query $Q\in \mathbb{R}^{B\times hs\times N\times1}$ and key $K\in \mathbb{R}^{B \times hs\times N\times1}$ are effective at capturing global similarities. The SCA operation is expressed as follows:
\begin{equation}
    \mathrm{{Q_i}}= \mathrm{W_i^Q({F_i})} \in {\mathbb{R}^{\mathrm{B \times hs \times {N_i} \times 1}}},
\end{equation}
\begin{equation}
    \mathrm{{K_i} = W_i^K({M_i})\in {\mathbb{R}^{B \times hs \times N_i^{m} \times 1}}},
\end{equation}
\begin{equation}
    \mathrm{{V_i} = W_i^V ({M_i}) \in {\mathbb{R}^{B \times hs \times N_i^{m} \times dim_{h}}}},
\end{equation}
\begin{equation}
    \mathrm{Attn\left( {{Q_i},{K_i}} \right) = Softmax (\frac{{{Q_i}K_i^T}}{{\sqrt {{d_k}} }}) \in {\mathbb{R}^{B \times hs \times {N_i} \times N_i^m}}},
\end{equation}
\begin{equation}
    \mathrm{{P}_i = Attn\left( {{Q_i},{K_i}} \right) V_i^T \in {\mathbb{R}^{B \times hs \times {N_i} \times dim_{h}}}},
\end{equation}
\begin{equation}
    \mathrm{SCA\left( {{F_i},{M_i}} \right) = W_i^O \left( Cat\left( {P{_0},...,P{_{hs}}} \right)\right) \in {\mathbb{R}^{B\times{N_i}\times{C_i}}}},
\end{equation}
where $W_i^Q$, $W_i^K$ and $W_i^V$ are transformation matrices used to map features. $B$ denotes the batch size, $hs$ represents the number of attention heads, $N$ is the number of tokens, and $dim_{h}$ is the dimension of each head.

\subsection{Local Perception Module (LPM)}
Global attention excels at capturing long-range dependencies, but it often overlooks local context. To address the lack of local perception in standard self-attention and cross-attention mechanisms, we introduce a Local Perception Module (LPM) in the CLB, drawing inspiration from backbones such as XCiT~\cite{ali2021xcit}, SMT~\cite{lin2023scale}, and CMT~\cite{guo2022cmt}. As shown in Fig.~\ref{lpm}, the LPM can be derived using the following equation:
\begin{equation}
    \mathrm{{x_d} = DWCon{v_{3 \times 3}}\left( {\sigma \left( {DWCon{v_{1 \times 1}}\left( x \right)} \right)} \right)},
\end{equation}
\begin{equation}
    \mathrm{\omega  = Sigmoid\left( {f^{Li}(\sigma \left( {f^{Li}\left( {AvgPool\left( {{x_d}} \right)} \right)} \right)} \right)},
\end{equation}
\begin{equation}
    \mathrm{y = x + DWCon{v_{1 \times 1}}\left( {\omega  \odot {x_d}} \right)},
\end{equation}
where $DWConv$ denotes a depthwise separable convolution, $\sigma$ represents the ReLU activation function, $f^{Li}$ is the $nn.Linear$ operation, and $\odot$ symbolizes matrix multiplication by channel.

\begin{table*}[!t]
\caption{Performance comparisons with SOTA medium-weight models on ADE20K~\cite{zhou2017scene} and Cityscapes~\cite{cordts2016cityscapes}.}
\begin{center}	
\scalebox{1.0}{
\begin{tabular}{l||lcc|cccc}
\toprule
\multirow{2}{*}{\textbf{Method}} & \multirow{2}{*}{\textbf{Year}} & \multirow{2}{*}{\textbf{Backbone}} & \multirow{2}{*}{\textbf{Params. (M)}$\downarrow$} & \multicolumn{2}{c}{\textbf{ADE20K}~\cite{zhou2017scene}} & \multicolumn{2}{c}{\textbf{Cityscapes}~\cite{cordts2016cityscapes}} \\
&   &   &   & \textbf{GFLOPs}$\downarrow$  & \textbf{mIoU (\%)}$\uparrow$  & \textbf{GFLOPs}$\downarrow$  & \textbf{mIoU (\%)}$\uparrow$      \\ \midrule

CCNet~\cite{huang2019ccnet}   & 2019 ICCV   & ResNet-101  & 68.9  & 278.4  & 43.7   & 2224.8   & 79.5    \\
EncNet~\cite{zhang2018context}  & 2018 CVPR  &  ResNet-101 & 55.1 
& 218.8 & 44.7 & 1748.0 & 76.9\\
DeepLab-V3+~\cite{chen2018encoder}& 2018 ECCV & ResNet-101  & 52.7  & 255.1 & 44.1 & 2032.3  & 80.9  \\
Mask2Former~\cite{cheng2022masked}   & 2022 CVPR  & ResNet-101 & 63.0 & 90.0 & 47.8  & - & -   \\
Auto-DeepLab~\cite{liu2019auto}    & 2019 CVPR & Auto-DeepLab-L  & 44.4   & -  & -   & 695.0  & 80.3   \\
OCRNet~\cite{yuan2020object}     & 2020 ECCV   & HRNet-W48   & 70.5  & 164.8  & 45.6   & 1296.8  & 81.1   \\

\midrule

SegFormer~\cite{xie2021segformer}   & 2021 NeurIPS  & MiT-B1 & 13.7 & 15.9  & 42.2   & 243.7  & 78.5  \\
SegDeformer~\cite{shi2022transformer} & 2022 ECCV & MiT-B1 & 14.4 & - & 44.1 
& - & - \\
SeaFormer~\cite{wanseaformer} & 2023 ICLR & SeaFormer-B & \textbf{8.6} & \textbf{1.8} & 40.2 & - & 77.7\\
SCTNet~\cite{xu2024sctnet} & 2024 AAAI & SCTNet-B & 17.4 & - & 43.0 & - & 79.8 \\
FeedFormer~\cite{shim2023feedformer}   & 2023 AAAI  & MiT-B1  & 17.3  & 20.7  & 44.2  
& 256.0  & 79.0   \\
U-MixFormer~\cite{yeom2025u}   & 2025 WACV  & MiT-B1  & 24.0  & 17.8   & 45.2 & 246.8  & 79.9  \\

SFNet & 2024 IJCV & ResNet-18 & \underline{12.3} & - & - &- & \underline{80.1} \\
MetaSeg~\cite{kang2024metaseg}  & 2024 WACV  & MiT-B1  & 16.0  & \underline{12.4}  & 43.8 & \textbf{219.0}  & 78.6 \\
PEM~\cite{cavagnero2024pem} & 2024 CVPR & ResNet-50 & 35.6 & 46.9 & \textbf{45.5} & \underline{240.0} & 79.9 \\
VWFormer~\cite{yan2024multiscale} & 2024 ICLR & MiT-B1 &13.7 & 13.2 & 43.2 & 289.0 & 79.0 \\
\midrule
\rowcolor{gray!20}SCASeg (Ours)  & \textbf{-}   & MiT-B1   & 23.4 & 17.4 & \underline{45.4}  & 248.8   & \textbf{80.3}\\ 
\midrule

SegNeXt~\cite{guo2022segnext}    & 2022 NeurIPS   & MSCAN-S  & \textbf{13.9} & \underline{15.9}  & 44.3 & \textbf{124.6}  & 81.3  \\
FeedFormer~\cite{shim2023feedformer}   & 2023 AAAI  & MSCAN-S  & 17.6 & 23.6  & 46.7  & 163.0  & 81.5   \\
MetaSeg~\cite{kang2024metaseg}  & 2024 WACV  & MSCAN-S  & 16.3  & \textbf{15.3}  & 45.9 & \underline{126.0}  & 81.3 \\
VWFormer~\cite{yan2024multiscale} & 2024 ICLR & MSCAN-S & \underline{15.5} & 22.5 & 46.2 & 196.0 & \textbf{81.7} \\
CPT~\cite{tang2025rethinking} & 2025 TIP & ResNet-101 & 48.5 & 100.0 & 46.8 & - & -\\
EDAFormer~\cite{yu2024embedding} & 2024 ECCV & EDAFormer-B & 29.4 & 32.0 & \textbf{49.0} & 605.9 & \underline{81.6}\\
OffSeg~\cite{zhang2025revisiting} & 2025 arxiv & OffSeg-L & 26.4 & 17.1 & 48.5 & 143.4 & \underline{81.6}\\
\midrule
\rowcolor{gray!20}SCASeg (Ours)  & \textbf{-}   & MSCAN-S   & 23.7  & 20.3  & \underline{48.8}   & 155.6  & \underline{81.6}\\ 

\midrule

SegFormer~\cite{xie2021segformer}     & 2021 NeurIPS   & MiT-B2  & \underline{27.5}  & 62.4  & 46.5  & 717.1  & 81.0  \\
SegDeformer~\cite{shi2022transformer} & 2022 ECCV & MiT-B2 & 27.6 & - & 47.5 & -&-\\
FeedFormer~\cite{shim2023feedformer}   & 2023 AAAI   & MiT-B2   & 29.1   & 42.7  & 48.0 & 522.7  & 81.5  \\
MetaSeg~\cite{kang2024metaseg}  & 2024 WACV  & MiT-B2  & 27.8  & \textbf{25.2}  & 46.3 & \textbf{420.0}  & 81.2 \\
VWFormer~\cite{yan2024multiscale} & 2024 ICLR & MiT-B2
& \textbf{27.4} & 46.6 & \underline{48.1} & \underline{469.0} & \underline{81.7}\\
CPT-M~\cite{tang2025rethinking} & 2025 TIP & ResNet-101 & 50.4 & 113.0 & 47.0 & - & -\\
CPT-M~\cite{tang2025rethinking} & 2025 TIP & D-ResNet-101 & 50.4 & 258.0 & \underline{48.1} & - & -\\
\midrule
\rowcolor{gray!20}SCASeg (Ours)  & \textbf{-}   & MiT-B2   & 35.2 & \underline{39.6} & \textbf{48.3}  & 516.6  & \textbf{81.9}\\

\bottomrule
\end{tabular}
}
\label{medium}
\end{center}
\end{table*}

\begin{table*}[!t]
\caption{Performance comparisons with SOTA heavy-weight models on ADE20K~\cite{zhou2017scene} and Cityscapes~\cite{cordts2016cityscapes}.}
\begin{center}	
\scalebox{1.0}{
\begin{tabular}{l||lcc|cccc}
\toprule
\multirow{2}{*}{\textbf{Method}} & \multirow{2}{*}{\textbf{Year}} & \multirow{2}{*}{\textbf{Backbone}} & \multirow{2}{*}{\textbf{Params. (M)}$\downarrow$} & \multicolumn{2}{c}{\textbf{ADE20K}~\cite{zhou2017scene}} & \multicolumn{2}{c}{\textbf{Cityscapes}~\cite{cordts2016cityscapes}} \\
&   &   &   & \textbf{GFLOPs}$\downarrow$  & \textbf{mIoU (\%)$\uparrow$}  & \textbf{GFLOPs}$\downarrow$  & \textbf{mIoU (\%)$\uparrow$}      \\ \midrule
Seg-B-Mask/16~\cite{strudel2021segmenter} & 2021 ICCV   & ViT-Base  & 106.0  & -  & 48.5  & -  & -   \\
MaskFormer~\cite{cheng2021per} & 2021 NeurIPS   & Swin-S  & 63.0  & 79.0  & 49.8  & -  & -   \\
SETR~\cite{zheng2021rethinking} & 2021 CVPR   & ViT-Large  & 318.3  & -  & 50.2  & -  & 82.2   \\
\midrule
SegNeXt~\cite{guo2022segnext}     & 2022 NeurIPS  & MSCAN-B  & \textbf{27.6}  & 34.9  & \underline{48.5}  & 275.7  & 82.6   \\
FeedFormer~\cite{shim2023feedformer}   & 2023 AAAI   & MSCAN-B  & 30.5   & 36.8  & 48.3 & 269.0  & 82.1  \\
MetaSeg~\cite{kang2024metaseg}  & 2024 WACV & MSCAN-B   & 29.6  & \textbf{30.4}  &  \underline{48.5}  & \textbf{251.1}  & \underline{82.7} \\
VWFormer~\cite{yan2024multiscale} & 2024 ICLR & MSCAN-B &\underline{28.3} & 35.7 & 48.1 & 302.0 & 82.3 \\
\midrule
\rowcolor{gray!20}SCASeg (Ours)  & \textbf{-}   & MSCAN-B   & 36.5 & \underline{33.5} & \textbf{49.6}  & \underline{261.0}  & \textbf{83.0}\\
\midrule

ContrastiveSeg~\cite{wang2021exploring} & 2021 ICCV &  HRNetV2-W48 & -& -& -& -&  81.4\\
SegFormer~\cite{xie2021segformer}   & 2021 NeurIPS  & MiT-B3  & \textbf{47.3}   & 79.0   & 49.4 &962.9 & 81.7\\
FeedFormer~\cite{shim2023feedformer}   & 2023 AAAI   & MiT-B3  & 48.3   & \underline{47.2}  & 49.5 & 682.0  & 81.9  \\
MetaSeg~\cite{kang2024metaseg}  & 2024 WACV  & MiT-B3  & \underline{47.7}  & \textbf{41.8}  & 48.7 & \textbf{645.0}  & 81.8 \\
VWFormer~\cite{yan2024multiscale} & 2024 ICLR & MiT-B3 & \textbf{47.3} & 63.3 & \underline{49.6} & 715.0 & \underline{82.4}\\
\midrule
\rowcolor{gray!20}SCASeg (Ours)  & \textbf{-}   & MiT-B3   & 55.1 & 56.3  & \textbf{50.1}  & \underline{675.0}  &\textbf{83.0}\\
\midrule
CLUSTSEG~\cite{liang2023clustseg} & 2023 ICML & ResNet-50 & - & - & 50.5 & - & -\\
SegFormer~\cite{xie2021segformer}   & 2021 NeurIPS          & MiT-B4  & 64.1       & 95.7   & 50.3 &1240.6 &81.9\\
FeedFormer~\cite{shim2023feedformer}   & 2023 AAAI   & MiT-B4  & 65.0   & \underline{63.8}  & 50.7 & 960.0  & 82.6  \\
MetaSeg~\cite{kang2024metaseg}  & 2024 WACV  & MiT-B4  & \textbf{63.6}  & \textbf{55.5}  & 50.5 & \textbf{923.0}  & 82.1 \\
VWFormer~\cite{yan2024multiscale} & 2024 ICLR & MiT-B4 & \underline{64.0} & 79.9 & \underline{50.8} & 993.0 & \underline{82.7}\\
MacFormer~\cite{xu2024macformer}   & 2024 arxiv   & MiT-B4   & 82.0 & 76.7  & \textbf{50.9}   & -  &-\\
\midrule
\rowcolor{gray!20}SCASeg (Ours)  & \textbf{-}   & MiT-B4   & 71.8 & 72.9  & \textbf{50.9}   & \underline{953.0}  &\textbf{83.2}\\
\midrule

ContrastiveSeg~\cite{wang2021exploring} & 2021 ICCV & OCR & -& -& -& -&  \underline{83.2}\\
SegFormer~\cite{xie2021segformer}   & 2021 NeurIPS          & MiT-B5  & \underline{84.7}       & 183.3  & 51.0 &1460.4&82.4\\
FeedFormer~\cite{shim2023feedformer}   & 2023 AAAI   & MiT-B5  & 85.6   & \underline{79.8}  & 51.2 & 1180.0  & 82.7  \\
 U-MixFormer~\cite{yeom2025u} & 2025 WACV  & MiT-B5  & 93.0  & 149.5  & 51.9 & \underline{1171.0} &83.1\\

ViT-CoMer~\cite{xia2024vit}  & 2024 CVPR & ViT-CoMer-B & 144.7    & -     & 48.8 &- &- \\
MetaSeg~\cite{kang2024metaseg}  & 2024 WACV  & MiT-B5  & 85.0  & \textbf{74.5}  & 51.4 & \textbf{1143.0}  & 82.5 \\
VWFormer~\cite{yan2024multiscale} & 2024 ICLR & MiT-B5 & \textbf{84.6} & 96.1 & 52.0 & 1213.0 & 82.8\\
MacFormer  & 2024 arxiv   & MiT-B5   & 103.0  & 152.4 & \textbf{52.8}  & -  &-\\
\midrule
\rowcolor{gray!20}SCASeg (Ours)  & \textbf{-}   & MiT-B5   & 92.4  & 88.9 & \underline{52.7}   & 1173.0  &\textbf{83.5}\\
\bottomrule
\end{tabular}
}
\label{heavy}
\end{center}
\end{table*}

\begin{table*}[!t]
\caption{Performance comparisons with SOTA models on COCO-Stuff 164k~\cite{caesar2018coco} and Pascal VOC2012~\cite{hoiem2009pascal}.}
\begin{center}
\begin{tabular}{l||lccc|cccc}
\toprule
\multirow{2}{*}{\textbf{Method}} & \multirow{2}{*}{\textbf{Year}} & \multirow{2}{*}{\textbf{Backbone}} & \multirow{2}{*}{\textbf{Params. (M)}$\downarrow$} & \multirow{2}{*}{\textbf{FPS (img/s)}$\uparrow$} & \multicolumn{2}{c}{\textbf{COCO-Stuff 164k}~\cite{caesar2018coco}} & \multicolumn{2}{c}{\textbf{Pascal VOC2012}~\cite{hoiem2009pascal}} \\
&      &      &    &       & \textbf{GFLOPs}$\downarrow$         & \textbf{mIoU (\%)$\uparrow$ }        & \textbf{GFLOPs}$\downarrow$       & \textbf{mIoU (\%)$\uparrow$}       \\ \midrule
SegFormer~\cite{xie2021segformer}& 2021 NeurIPS  & MiT-B0 &  \underline{3.8}  &  \textbf{43.65} & 8.4  & 35.63 &  8.4 & 66.49 \\
FeedFormer~\cite{shim2023feedformer} & 2023 AAAI  & MiT-B0  & 4.5 & 34.80 & 7.8  & 39.03  &  7.8  & 68.49 \\
U-MixFormer~\cite{yeom2025u}& 2025 WACV & MiT-B0 & 6.1 & 38.94 & 6.1 & \underline{40.24} & 6.1 & \underline{71.16}  \\
MetaSeg~\cite{kang2024metaseg}  & 2024 WACV & MiT-B0 & 4.1 &  \underline{42.64} &  \textbf{3.9}  &  38.25  &  \textbf{3.9}   & 68.72  \\
VWFormer~\cite{yan2024multiscale}  & 2024 ICLR & MiT-B0 &  \textbf{3.7}   & 39.63  &  \underline{5.8} &  36.28  &  \underline{5.8}   & 70.58\\
\midrule
\rowcolor{gray!20}SCASeg (Ours)  & \textbf{-}   & MiT-B0   & 6.0 & 40.25 & 5.9 &  \textbf{40.56}   & 5.9  &  \textbf{72.35} \\
\midrule

SegNeXt~\cite{guo2022segnext}   & 2022 NeurIPS & MSCAN-T &  \textbf{4.3} &  \textbf{33.16} &  \underline{6.6} & 38.70  &  \underline{6.6} & 76.27   \\
FeedFormer~\cite{shim2023feedformer} & 2023 AAAI  & MSCAN-T  & 5.0 & 25.98 & 9.3 & 39.39   & 9.3  & 74.80 \\
U-MixFormer~\cite{yeom2025u}& 2025 WACV & MSCAN-T & 6.7 & 27.69 & 7.6 & \underline{40.04}   & 7.6  & \underline{77.37}  \\
MetaSeg~\cite{kang2024metaseg}  & 2024 WACV & MSCAN-T  &  \underline{4.7} &  \underline{31.74} &  \textbf{5.5} & 39.70 &  \textbf{5.5} & 74.98  \\
VWFormer~\cite{yan2024multiscale}   & 2024 ICLR & MSCAN-T & 5.8 & 28.06 & 13.6 &  38.85  & 13.6   & 76.53  \\
\midrule
\rowcolor{gray!20}SCASeg (Ours)  & \textbf{-}   & MSCAN-T   & 6.5 & 29.04 & 7.4 & \textbf{40.89}   & 7.4  &  \textbf{77.88}  \\

\midrule

 FeedFormer~\cite{shim2023feedformer} &  2023 AAAI  &   SegMAN-T  &   \underline{3.1} &   41.66 &   7.9 &   41.54   &   7.9  &   75.65 \\
U-MixFormer~\cite{yeom2025u}& 2025 WACV &   SegMAN-T & 4.3 & 41.93 & 7.0 & \underline{41.98}  & 7.0  &  \underline{75.87}  \\
 MetaSeg~\cite{kang2024metaseg}  &  2024 WACV &   SegMAN-T  &  2.9 &  42.81 &  \textbf{4.5} &  41.68 &  \textbf{4.5} &  75.70  \\
 VWFormer~\cite{yan2024multiscale}   &  2024 ICLR &   SegMAN-T &  \textbf{2.8} &   42.15 &  6.3 &    41.38  &  6.3   &   75.26  \\
 SegMAN~\cite{fu2025segman}   &  2025 CVPR &   SegMAN-T &  6.4 &  \textbf{44.52} &  \underline{6.2} &  41.30  &  \underline{6.2} &  75.10   \\
\midrule
\rowcolor{gray!20} SCASeg (Ours)  &  \textbf{-}   &  SegMAN-T   &  4.3 &  \underline{43.02} &  6.9 & \textbf{42.22}   &  6.9  &  \textbf{76.02}  \\

\midrule

SegNeXt~\cite{guo2022segnext}   & 2022 NeurIPS & MSCAN-S &  \textbf{13.9} &  \textbf{30.91} &  \underline{15.9} & 41.42  &  \underline{15.9} & 78.62 \\
FeedFormer~\cite{shim2023feedformer} & 2023 AAAI  & MSCAN-S  & 17.6 & 24.16 & 23.6 & 42.61   & 23.6  & 77.42 \\
U-MixFormer~\cite{yeom2025u}& 2025 WACV & MSCAN-S & 24.3 & 26.55 & 20.8 & \underline{42.91}   & 20.8  & \underline{79.23}   \\
MetaSeg~\cite{kang2024metaseg}  & 2024 WACV & MSCAN-S  & 16.3  &  \underline{29.83} &  \textbf{15.3} & 42.13 &  \textbf{15.3} & 76.73  \\
VWFormer~\cite{yan2024multiscale}   & 2024 ICLR & MSCAN-S &  \underline{15.5} & 27.37  & 22.5 &  41.76  & 22.5   & 78.95 \\
\midrule
\rowcolor{gray!20}SCASeg (Ours)  & \textbf{-}   & MSCAN-S   & 23.7 & 27.71 & 20.3 & \textbf{43.65}   & 20.3  & \textbf{79.48} \\

\midrule

SegFormer~\cite{xie2021segformer}& 2021 NeurIPS  & MiT-B1  &  \underline{13.7} &  \textbf{31.44} & 15.9  & 40.97  &  15.9 & 71.13 \\
FeedFormer~\cite{shim2023feedformer} & 2023 AAAI  & MiT-B1  & 17.3 & 22.29 & 20.7 & 42.42  &  20.7   & 71.77 \\
U-MixFormer~\cite{yeom2025u}  & 2025 WACV & MiT-B1 & 24.0  & 22.06 & 17.8  & \underline{42.71}  & 17.8 & \underline{74.40}  \\
MetaSeg~\cite{kang2024metaseg}  & 2024 WACV & MiT-B1 &  \textbf{16.0} &  \underline{27.12} &  \textbf{12.4} & 42.04 &  \textbf{12.4}  & 73.30  \\
VWFormer~\cite{yan2024multiscale}   & 2024 ICLR & MiT-B1   &  \underline{13.7} & 23.87 &  \underline{13.2} & 41.54  &  \underline{13.2}   & 73.98  \\
\midrule
\rowcolor{gray!20}SCASeg (Ours)  & \textbf{-}   & MiT-B1   & 23.4 & 25.37 & 17.4 &  \textbf{43.19} & 17.4  & \textbf{75.24} \\

\midrule

 SegFormer~\cite{xie2021segformer}&  2021 NeurIPS   &  MiT-B2   &  \underline{27.5} &   \textbf{29.72} &  62.4 &   43.42   &  62.4   &   78.74  \\
 FeedFormer~\cite{shim2023feedformer} &  2023 AAAI  &  MiT-B2   &  29.1  &   27.56 &  42.7   &    44.79    &  42.7   &  78.90 \\
 U-MixFormer~\cite{yeom2025u}  &  2025 WACV &  MiT-B2  &   35.8 & 26.34 &  40.0    &    \underline{45.47}   &  40.0    &   \underline{79.43} \\
 MetaSeg~\cite{kang2024metaseg}  &  2024 WACV &  MiT-B2 &  27.8 &  \underline{28.47} &  \textbf{25.2}  &  44.50  &  \textbf{25.2}  & 77.76  \\
 VWFormer~\cite{yan2024multi}   &  2024 ICLR &  MiT-B2   &  \textbf{27.4} &  27.68 &  46.6     &   45.18      &  46.6     & 79.10  \\
\midrule
\rowcolor{gray!20} SCASeg (Ours)  &  \textbf{-}   &  MiT-B2   &  35.2 &  27.00 &  \underline{39.6} & \textbf{45.85}   &  \underline{39.6}   &  \textbf{80.15} \\



\bottomrule
\end{tabular}
\label{coco&pascal}
\end{center}
\end{table*}

\section{Experiments}
\subsection{Experimental Settings}

\subsubsection{Datasets}

We conducted experiments on four publicly available datasets: ADE20K~\cite{zhou2017scene}, Cityscapes~\cite{cordts2016cityscapes}, COCO-Stuff 164K~\cite{caesar2018coco}, and Pascal VOC2012~\cite{hoiem2009pascal}. ADE20K~\cite{zhou2017scene} contains 150 semantic categories, with 20,210 training images, 2,000 validation images, and 3,352 test images. Cityscapes~\cite{cordts2016cityscapes} focuses on urban scenes with 19 categories, including 2,975 training images, 500 validation images, and 1,525 test images. COCO-Stuff164K~\cite{caesar2018coco} has 164,000 images annotated with 171 categories, enhancing the COCO dataset with detailed scene parsing. Pascal VOC2012~\cite{hoiem2009pascal} includes 11,530 images across 20 categories, with pixel-level annotations for classification and segmentation tasks.

\subsubsection{Implementation Details}

Unless otherwise specified, all models are trained using the AdamW optimizer with an initial learning rate of 6e$-$5 for 160K iterations, following a polynomial learning rate decay schedule.
We utilized the \textit{mmsegmentation}~\cite{mmseg2020} codebase (Version 1.2.2) to train our model on eight NVIDIA A100 GPUs. In our experiments, we employed LVT, MiT-B0$\sim$5, and MSCAN as backbone networks, while keeping the encoder unchanged to ensure fair comparisons across different methods. Standard data augmentation strategies were applied, including random horizontal flipping, random scaling with ratios ranging from 0.5 to 2.0, and random cropping to $512 \times 512$ for ADE20K~\cite{zhou2017scene}, COCO-Stuff 164k~\cite{caesar2018coco}, and Pascal VOC2012~\cite{hoiem2009pascal}, and $1024 \times 1024$ for Cityscapes~\cite{cordts2016cityscapes}. The batch size was set to 16 for ADE20K, COCO-Stuff 164k, and Pascal VOC2012, and 8 for Cityscapes. All results are reported using single-scale inference, and performance is evaluated using mean Intersection over Union (mIoU).



\subsection{Experimental Results}

We categorized SOTA models into lightweight, medium-weight, and heavyweight. The main results on ADE20K~\cite{zhou2017scene} and Cityscapes~\cite{cordts2016cityscapes} are shown in Tables~\ref{light},~\ref{medium}, and~\ref{heavy}. 
For COCO-Stuff 164k~\cite{caesar2018coco} and Pascal VOC2012~\cite{hoiem2009pascal}, we conducted fair comparisons using the same backbone networks, with the key results presented in Table~\ref{coco&pascal}.

\begin{figure*}[ht]
	\centerline{\includegraphics[width=18cm]{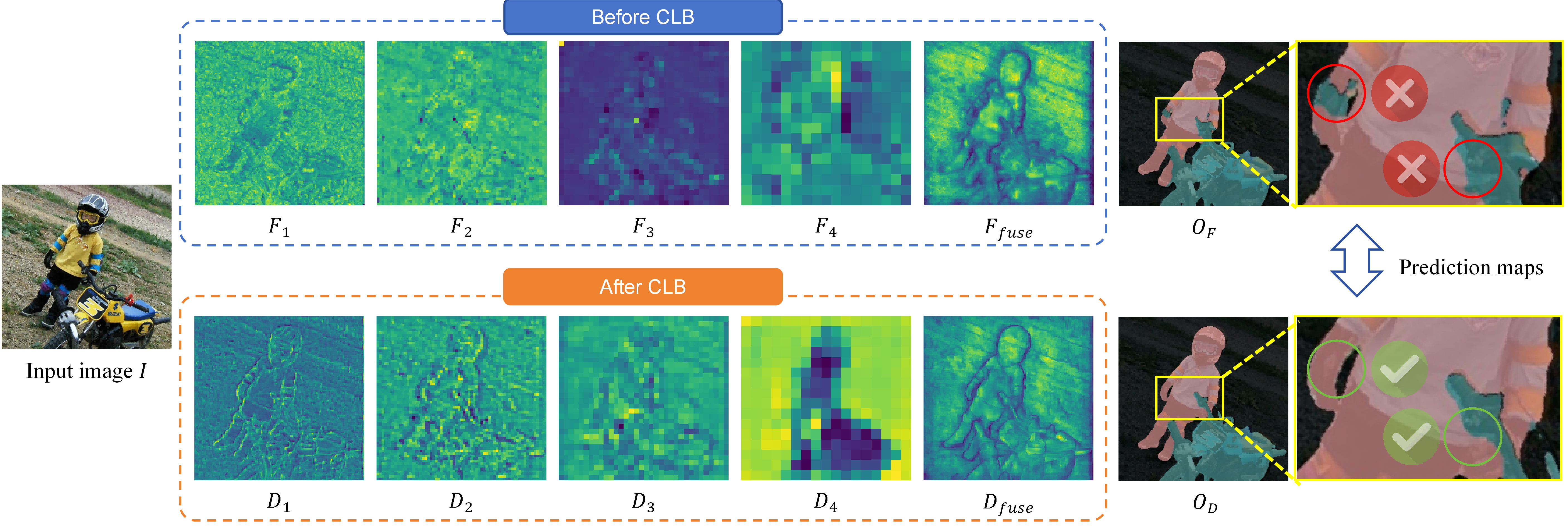}}
	\caption{Visualized feature heatmaps before and after introducing the Cross-Layer Block (CLB).}
	\label{featuremap}
    \vspace{-3mm}
\end{figure*}

\begin{figure}[ht]
	\centerline{\includegraphics[width=9cm]{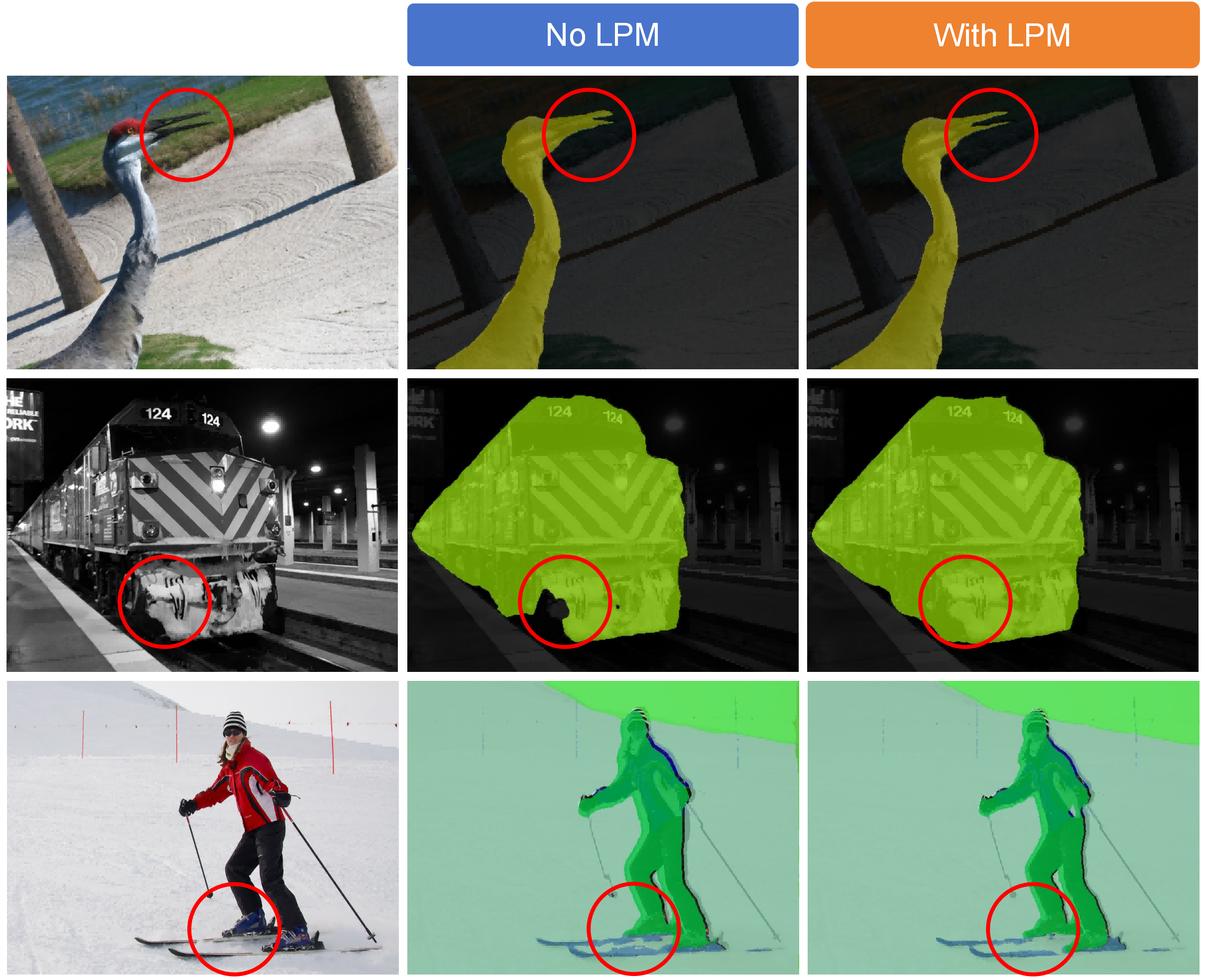}}
	\caption{Visual comparison of segmentation results with and without using our LPM.}
	\label{wlpmc}
    \vspace{-3mm}
\end{figure}

\noindent\textbf{ADE20K~\cite{zhou2017scene} \& Cityscapes~\cite{cordts2016cityscapes}.}
\textit{Lightweight models:} 
We presented the performance of the lightweight models in Table~\ref{light}. As indicated in the table, our lightweight model, SCASeg (MiT-B0), achieved an mIoU of 41.6\% on ADE20K~\cite{zhou2017scene}, utilizing only 6.0 million parameters and 5.9 GFLOPs. In comparison to SegFormer~\cite{xie2021segformer} (MiT-B0), SCASeg (MiT-B0) achieves a 4.2\% improvement in mIoU while reducing computational cost by 29.7\%. Although SDPT-Tiny~\cite{cao2024sdpt} and VWFormer~\cite{yan2024multiscale} (MiT-B0) have a slight advantage in parameter count, their GFLOPs are nearly the same as ours, and their mIoU is at least 2.2\% lower than that of our method. For Cityscapes~\cite{cordts2016cityscapes}, the performance difference becomes more evident, with our model achieving an mIoU of 79.3\% at just 101.7 GFLOPs. This represents a 3.1\% improvement in mIoU and an 18.9\% reduction in computational cost compared to SegFormer~\cite{xie2021segformer} (MiT-B0). Similarly, with LVT and MSCAN-T as backbones, our approach consistently achieves near-SOTA performance.

\begin{figure*}[htbp]
	\centerline{\includegraphics[width=17cm]{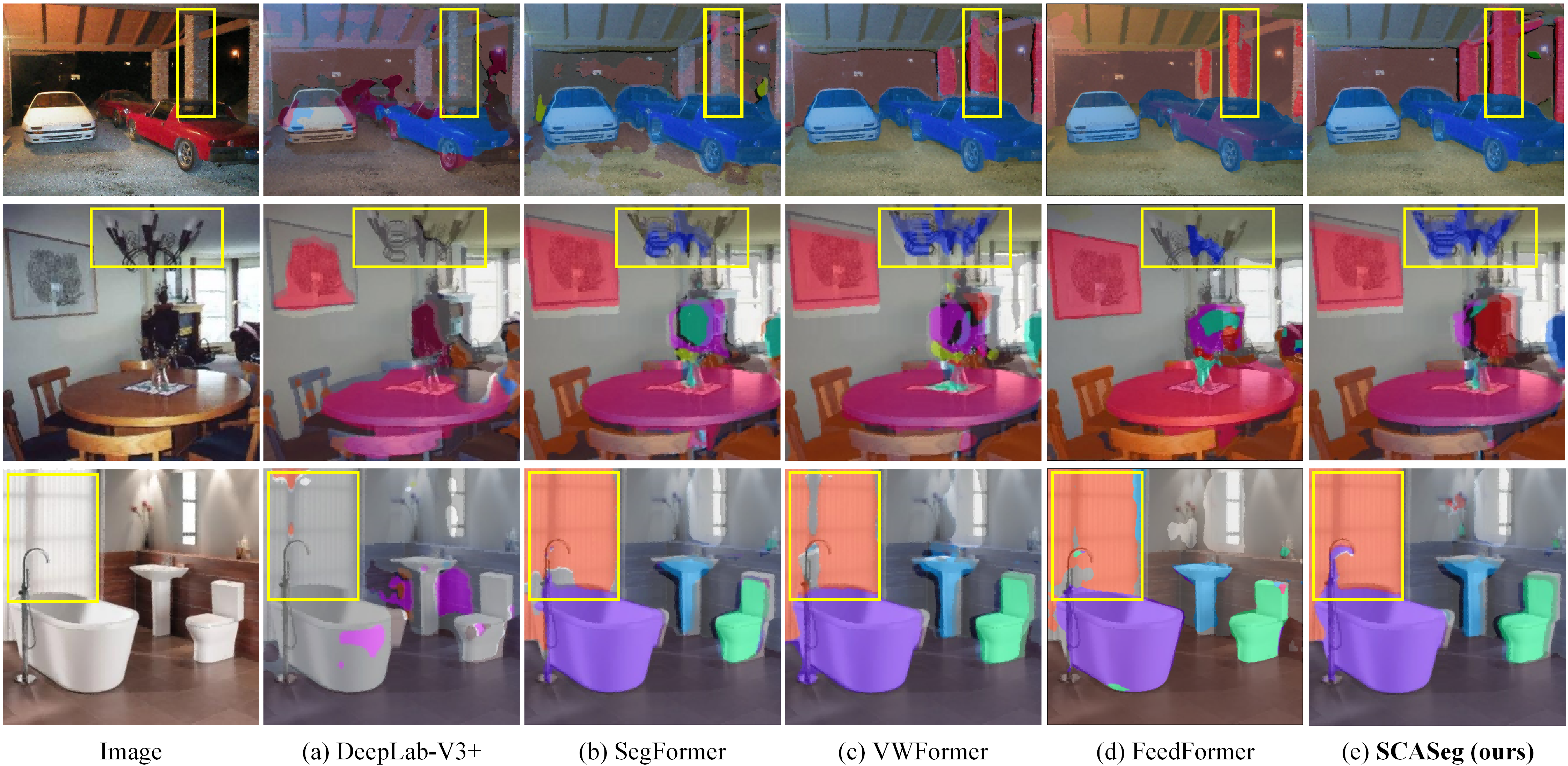}}
	\caption{Visual segmentation results obtained on ADE20K~\cite{zhou2017scene}.}
    \vspace{-3em}
	\label{ade20k}
\end{figure*}

\begin{figure*}[htbp]
	\centerline{\includegraphics[width=17cm]{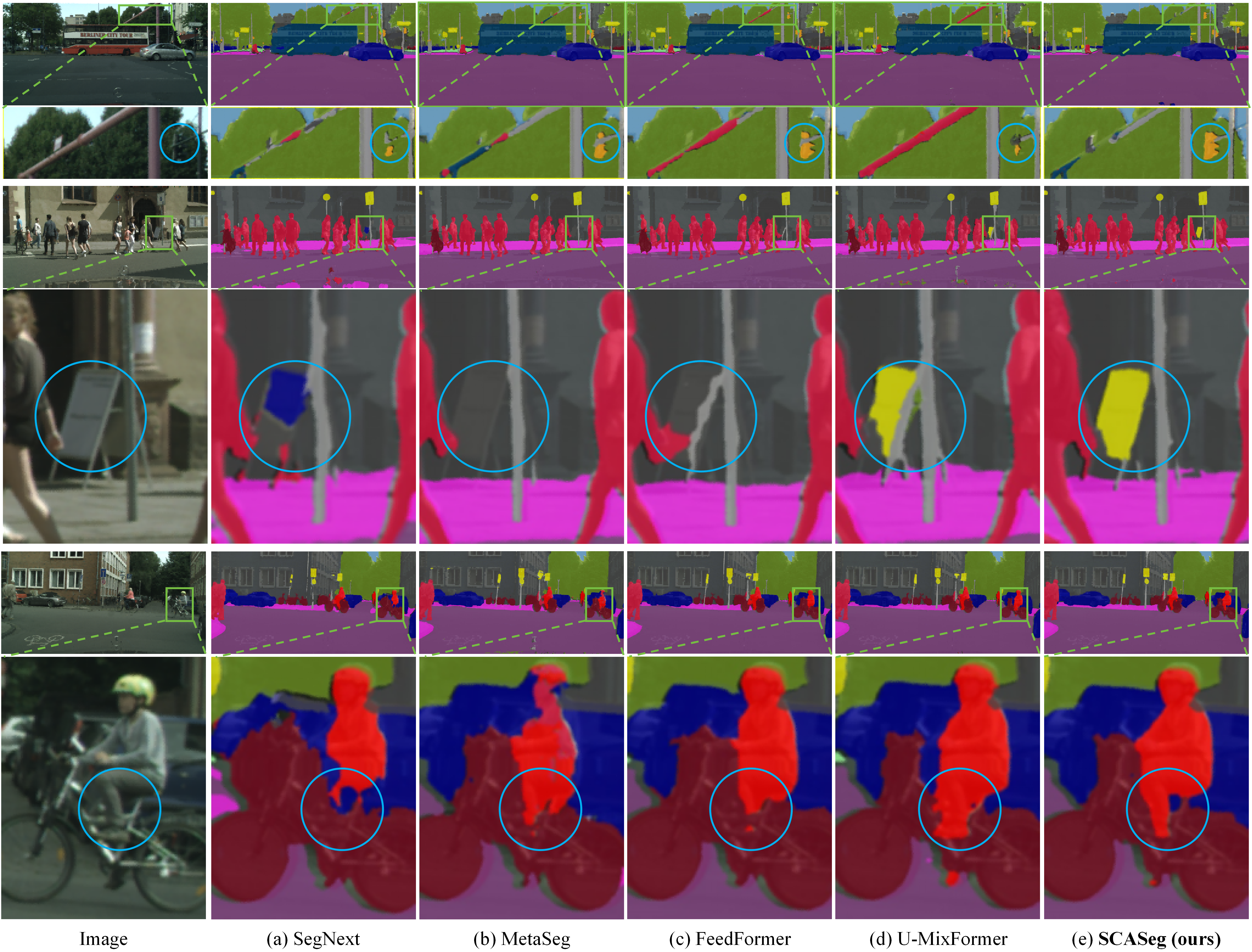}}
	\caption{Visual segmentation results obtained on Cityscapes~\cite{cordts2016cityscapes}.}
    \vspace{-1.5em}
	\label{cityscapes}
\end{figure*}

\begin{figure}[htbp]
	\centerline{\includegraphics[width=9cm]{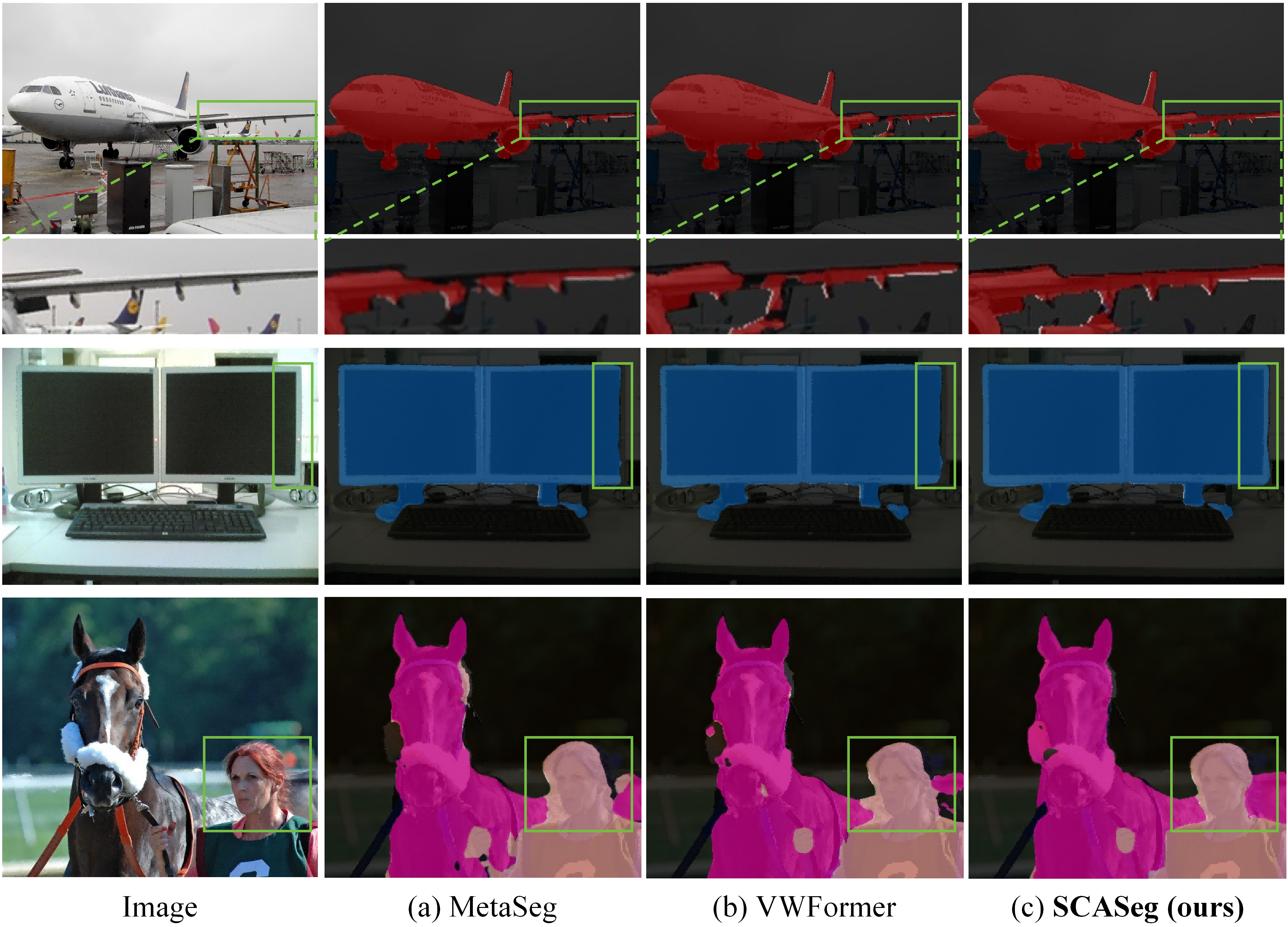}}
	\caption{Visual segmentation results obtained on Pascal VOC2012~\cite{hoiem2009pascal}.}
	\label{pascalvoc}
    \vspace{-3mm}
\end{figure}

\begin{figure}[htbp]
	\centerline{\includegraphics[width=9cm]{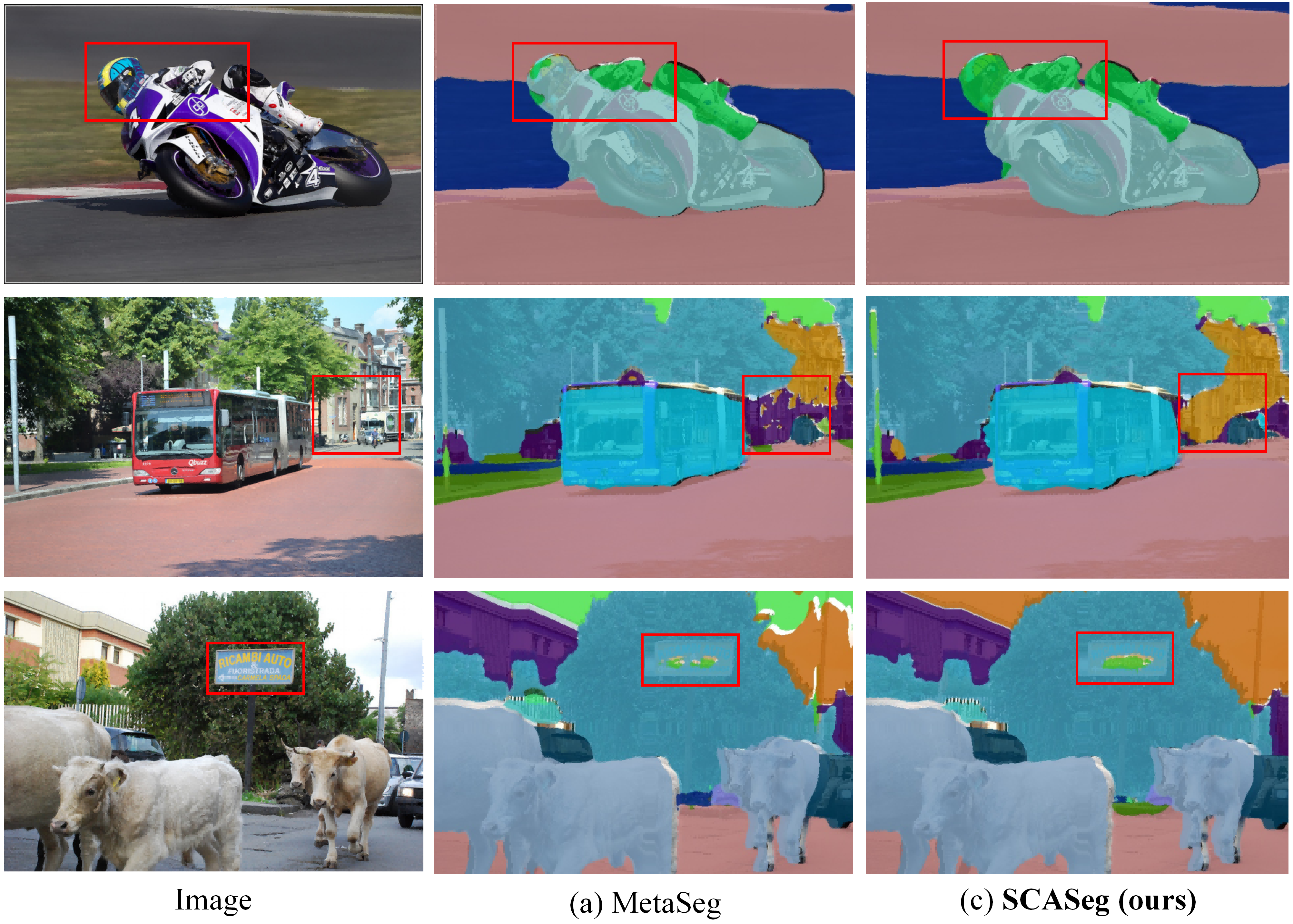}}
	\caption{Visual segmentation results obtained on COCO-Stuff 164K~\cite{caesar2018coco}.}
	\label{cocostuff164k}
    \vspace{-3mm}
\end{figure}

\textit{Medium-weight models:} 
As shown in Table~\ref{medium}, our SCASeg demonstrates superior performance compared to other methods when paired with equivalent heavy encoders. 
SCASeg (MiT-B1) achieved 45.4\% mIoU on ADE20K~\cite{zhou2017scene} with just 23.4 million parameters and 17.4 GFLOPs. Using MiT-B1 as the backbone, our approach reduces GFLOPs by 15.9\% 
while improving performance by 1.2\% compared to FeedFormer~\cite{shim2023feedformer}. The PEM model~\cite{cavagnero2024pem}, which achieves comparable 
performance (45.5\% vs. 45.4\%), requires 52\% more parameters and 2.7 times the GFLOPs (46.9 vs. 17.4) of our method. Although EfficientMod~\cite{ma2024efficient} achieves an mIoU that is 0.6\% higher than ours, it incurs an additional computational cost of nearly 11 GFLOPs. On the Cityscapes~\cite{cordts2016cityscapes} dataset, our approach achieves a new SOTA accuracy of 80.3\%. Similarly, when evaluated using MSCAN-S and MiT-B2 as backbones, SCASeg effectively balances performance and efficiency.

\textit{Heavyweight models:}
Table~\ref{heavy} presents the experimental comparison using heavy backbones, specifically MSCAN-B and MiT-B3/4/5. Our method also demonstrates solid results. For instance, on the ADE20K~\cite{zhou2017scene} dataset, SCASeg (MSCAN-B) achieves 49.6\% mIoU with only 33.5 GFLOPs. In comparison, VWFormer~\cite{yan2024multiscale} shows worse performance while with more computation cost (48.1\% mIoU with 35.7 GFLOPs). For the Cityscapes~\cite{cordts2016cityscapes} dataset, using MiT-B5 as the backbone, our method achieved an SOTA mIoU of 83.5\%, with a relatively small and justifiable cost of 1173.0 GFLOPs. Although the GFLOPs of MetaSeg~\cite{kang2024metaseg} are 2.6\% lower than ours (1143 vs. 1173), its mIoU is 1\% lower, indicating a clear performance difference. These experimental results demonstrate that, under the same conditions, our method strikes a better balance between performance and efficiency compared to other approaches, highlighting its distinct advantages and validating its effectiveness.

\noindent
\textbf{COCO-Stuff 164k~\cite{caesar2018coco} \& Pascal VOC2012~\cite{hoiem2009pascal}:}
In Table~\ref{coco&pascal}, we compared our SCASeg model with previous methods on the COCO-Stuff 164k~\cite{caesar2018coco} and Pascal VOC2012~\cite{hoiem2009pascal} datasets. To ensure a fair assessment, we selected different backbone configurations (MiT-B0/B1/B2, MSCAN-T/S, and SegMAN-T) and applied different methods under the same experimental conditions: 160k iterations with 8 GPUs, each processing a batch size of 8. The inference time per image was tested on a V100 GPU. As observed in Table~\ref{coco&pascal}, our method demonstrates outstanding performance compared with other methods.

\subsection{Visualization Results}

\noindent
\textbf{Visual Comparison of Feature Maps Before and After Applying CLB:}
Fig.~\ref{featuremap} presents a visual comparison of feature maps in the decoder before and after introducing the CLB. Before applying the CLB, features from different stages (F1–F4) lacked interaction, resulting in no exchange or complementary information between them. This limitation led to errors and inaccuracies in the segmentation results after direct fusion. However, after incorporating the CLB, it is evident that object boundaries are clearly visible at all stages, and the network demonstrates enhanced edge perception and class distinction, ultimately yielding more accurate segmentation outcomes.

\noindent
\textbf{Visual Comparison of Segmentation Results with and without Using LPM:} 
The Local Perception Module (LPM) is a critical component of our design. While attention mechanisms typically emphasize global context, they often overlook local perception. By incorporating the LPM, we address this limitation. As illustrated in Fig.~\ref{wlpmc}, there is a noticeable difference in the continuity of local information with and without the LPM. For instance, objects such as the long beak of a red-crowned crane or a snowboard are prone to misprediction due to the influence of surrounding larger objects. The LPM effectively alleviated this issue, enabling more consistent segmentation of small details.

\textbf{Visual Comparison of Segmentation Results:}
Figs.~\ref{ade20k},~\ref{cityscapes},~\ref{pascalvoc}, and~\ref{cocostuff164k} show the visual comparison of the segmentation results obtained on the ADE20K~\cite{zhou2017scene}, Cityscapes~\cite{cordts2016cityscapes}, Pascal VOC2012~\cite{hoiem2009pascal}, and COCO-Stuff 164k~\cite{caesar2018coco} datasets, respectively, using our SCASeg and SOTA methods. The highlighted areas indicate regions where SCASeg outperforms the other methods in segmentation quality. This improvement is evident in two main aspects: first, the prediction accuracy for objects within the same category has increased (\textit{e.g.}, the pole next to the car and the chandelier); second, boundary segmentation accuracy has improved (\textit{e.g.}, billboards). Additionally, small objects, such as traffic lights, are accurately detected and predicted. Compared to SOTA methods, SCASeg achieves better recognition of object details near boundaries. This indicates that our model captures a more relevant visual context by leveraging the capacity of the CLB decoder strategy.

\subsection{Ablation Studies}

\textbf{Effectiveness of Strip Cross-Attention (SCA):}
In Table~\ref{att}, we demonstrate the effectiveness of incorporating SCA within the decoder. SCA serves as a core component of the CLB, primarily facilitating cross-level feature enhancement and interaction among different hierarchical layers. 
As shown in this table, incorporating Self-Attention in the Decoder stage leads to a 3.75\% improvement in segmentation accuracy. Compared to Self-Attention (SA), Strip Cross-Attention (SCA) achieves a notable reduction in computational overhead, lowering FLOPS by 38.9\%, while still delivering a 1.47\% improvement in mIoU. Comparisons with Cross-Attention (CA) and Strip Cross-Attention (SCA) show that Strip Cross-Attention not only boosts performance but also reduces parameter count by 0.3M and computation by 0.4G, with a slight advantage in inference speed as well. This demonstrates that SCA not only preserves high segmentation performance but also significantly enhances computational efficiency, making it a highly effective alternative for applications with limited resources.

\noindent
\textbf{Effectiveness of the Local Perception Module (LPM):}
Table~\ref{att} presents the results of combining SCA with LPM, forming the complete CLB structure. With the addition of LPM, parameter count and computational load become comparable to those of Cross-Attention (CA), yet this combination achieves an increase in segmentation performance of over 0.5\%. Moreover, inference speed remains nearly identical. When combined with LPM, CA also yields strong results—improving over CA alone by 0.32\% and 0.63\% on MiT-B0 and MSCAN-T, respectively—demonstrating LPM’s effectiveness. However, CA+LPM introduces higher complexity (6.4M parameters, 6.3G FLOPs) than SCA+LPM (6.0M parameters, 5.9G FLOPs), and its accuracy is still 0.28\% lower than that of SCA+LPM. In Table~\ref{lpm2}, we conducted a comparative experiment, using SENet~\cite{hu2018squeeze} within LPM to enhance the model's sensitivity to local information. When compared with other channel attention mechanisms (CBAM~\cite{woo2018cbam}, ECANet~\cite{wang2020eca}, CooAtt~\cite{hou2021coordinate}), SENet~\cite{hu2018squeeze} achieved the best segmentation accuracy, outperforming CooAtt by more than 0.3\%. SENet's integration with LPM significantly strengthens the model’s local modeling capability, enhancing coherence in segmentation and reducing misclassification for small objects or local regions of the same class.

\begin{table}[!t]
\caption{Ablation studies of SCA on Pascal VOC2012~\cite{hoiem2009pascal}. SA: Self-Attention, CA: Cross-Attention, SCA: Strip Cross-Attention, LPM: Local Perception Module.}
\begin{center}
\scalebox{0.95}{
\begin{tabular}{l||cccc}
\toprule
\textbf{Method} & \textbf{Params.}$\downarrow$ & \textbf{GFLOPs}$\downarrow$ & \textbf{mIoU} (\%)$\uparrow$ & \textbf{FPS (img/s)}$\uparrow$ \\
\midrule
MiT-B0 & 3.8 M & 8.4 & 66.49 & 44.0 \\ \midrule
+ SA  & 4.3 M  & 9.0  & 70.06 (\textcolor{green}{+3.57}) & 45.0        \\
+ CA  & 6.0 M  & 5.9  & 71.79 (\textcolor{green}{+5.30}) & 42.0        \\
 + CA + LPM  &   6.4 M &   6.3 &   72.11 (\textcolor{green}{+5.62}) &   41.0     \\
+ SCA  & 5.7 M & 5.5  & 71.53 (\textcolor{green}{+5.04})  & 43.0 \\ 
\rowcolor{gray!20}+ SCA + LPM & 6.0 M & 5.9    & 72.39 (\textcolor{green}{+5.90}) & 42.0\\ \midrule

MSCAN-T & 4.3 M & 6.6 & 76.27 & 33.0 \\ \midrule
+ SA  & 4.8 M  & 10.4 & 76.92 (\textcolor{green}{+0.65})& 30.0        \\
+ CA  & 6.7 M  & 7.3  & 77.38 (\textcolor{green}{+1.11})& 29.0        \\
  + CA + LPM   &  6.9 M &  7.6 &  78.01 (\textcolor{green}{+1.74}) &  27.0     \\
+ SCA & 6.3 M & 6.9   & 76.74 (\textcolor{green}{+0.47})& 31.0\\ 
\rowcolor{gray!20}+ SCA + LPM & 6.5 M    & 7.4    & 77.88 (\textcolor{green}{+1.61})& 29.0 \\

\bottomrule
\end{tabular}
}
\label{att}
\end{center}
\end{table}

\begin{table}[!t]
\caption{Ablation studies of the Local Perception Module (LPM) on Pascal VOC2012~\cite{hoiem2009pascal}.}
\begin{center}
\begin{tabular}{l||ccc}
\toprule
\textbf{Method} & \textbf{Params. (M)}$\downarrow$ & \textbf{GFLOPs}$\downarrow$ & \textbf{mIoU (\%)$\uparrow$} \\ \midrule
SCA (MiT-B0) & 5.7  & 5.5    & 71.53 \\ \midrule
+ LPM (CBAM)   & 5.9    & 5.7   & 71.29 (\textcolor{red}{-0.24})  \\ 
+ LPM (ECANet) & 5.9    & 5.7   & 71.58 (\textcolor{green}{+0.05}) \\ 
+ LPM (CooAtt)   & 6.0    & 5.7   & 71.87 (\textcolor{green}{+0.34})\\ 
\rowcolor{gray!20}+ LPM (SENet)  & 6.0    & 5.9   & 72.39 (\textcolor{green}{+0.86})\\
\midrule
SCA (MSCAN-T) & 6.3  & 6.9  & 76.74  \\ \midrule
+ LPM (CBAM)   & 6.5  & 7.1   & 76.79 (\textcolor{green}{+0.05}) \\ 
+ LPM (ECANet) & 6.5  & 7.1   & 76.49 (\textcolor{red}{-0.25}) \\ 
+ LPM (CooAtt)   & 6.5   & 7.1   & 77.57 (\textcolor{green}{+0.83})  \\ 
\rowcolor{gray!20}+ LPM (SENet)  & 6.5    & 7.4   & 77.88 (\textcolor{green}{+1.14}) \\
\bottomrule
\end{tabular}
\label{lpm2}
\end{center}
\end{table}

\noindent
\textbf{Effect of Different Channel Dimensions of Key and Value:}

To investigate the effect of compressing the channel dimensions of the query and key in our SCA module, we conducted an ablation study on the PASCAL VOC2012 dataset using MSCAN-T as the backbone. Specifically, we varied the output dimensions of the query ($Q$) and key ($K$) projections (shared across heads) while keeping the value projection and all other settings unchanged.
As shown in Table~\ref{dim}, reducing the query/key dimension to 1 channel achieves a robust mIoU of 77.88\%, with the lowest parameter count (6.5M), the lowest computational cost (7.4 GFLOPs), and the highest inference speed (29.0 FPS). Increasing the dimension from 1 to 8 slightly improves the mIoU to 77.99\%, suggesting that a wider $Q/K$ space may capture more fine-grained attention. However, the performance gain plateaus quickly, while model size and complexity rise steadily.
Overall, the 1-channel setting offers an optimal balance of efficiency and accuracy, with the attention computation retaining strong semantic expressiveness due to the uncompressed value features. This validates our design of using compressed $Q/K$ with high-dimensional $V$ to achieve compact yet effective attention maps.

\noindent
\textbf{Effectiveness of the Cross-Layer Strategy: }
The comparative experiments on cross-layer strategies highlight the necessity of information exchange across features at different stages. During the encoding phase, four processing stages generate features that vary in semantic richness and detail when passed to the decoder. Shallow stages, such as Stage 1 and Stage 2, retain more detailed information as they undergo less downsampling and fewer convolutional operations, resulting in clearer edge information. In contrast, deeper stages, such as Stage 3 and Stage 4, capture richer contextual information due to more extensive feature abstraction. This difference is visually evident in Fig.~\ref{featuremap}, which illustrates the characteristics of the feature maps. The cross-layer strategy facilitates mutual feature enhancement across stages, allowing each stage to compensate for the limitations of the others. 
As shown in Table~\ref{crosslayer}, applying the CLB module across all four stages and enabling cross-layer operations significantly boosts segmentation performance (72.35\% for MiT-B0, 77.88\% for MSCAN-T). This improvement underscores the effectiveness and necessity of cross-layer operations in achieving high segmentation accuracy.

\begin{table}[t]
\caption{Ablation studies of Query and Key's Channel dimensions using the MSCAN-T backbone on Pascal VOC2012~\cite{hoiem2009pascal}.}
\begin{center}
\scalebox{0.95}{
\begin{tabular}{c||cccc}
\toprule
\textbf{Channels} & \textbf{Params. (M)$\downarrow$} & \textbf{GFLOPs$\downarrow$} & \textbf{mIoU (\%)$\uparrow$} & \textbf{FPS (img/s)$\uparrow$} \\ \midrule

\rowcolor{gray!20}1 & 6.5  & 7.4  & 77.88 & 29.0  \\ \midrule
2   & 6.6  & 7.5   & 77.90 & 29.0 \\ 
4 & 6.6  & 7.5   & 77.94  & 28.0 \\ 
8   & 6.7   & 7.6   & 77.99 & 28.0 \\ 
\bottomrule
\end{tabular}
}
\label{dim}
\end{center}
\end{table}

\begin{table}[t]
\caption{Ablation studies of Cross-Layer Block (CLB) on Pascal VOC2012~\cite{hoiem2009pascal}.}
\begin{center}
\scalebox{0.92}{
\begin{tabular}{l||cccc|ccc}
\toprule
\multirow{2}{*}{\textbf{Method}} & \multicolumn{4}{c|}{\textbf{Cross Layer}} & \multirow{2}{*}{\textbf{Params.$\downarrow$}} & \multirow{2}{*}{\textbf{FLOPs$\downarrow$}} & \multirow{2}{*}{\textbf{mIoU (\%)$\uparrow$}} \\
& 4      & 3      & 2     & 1     &    &     &         \\ \midrule
\multirow{4}{*}{\begin{tabular}[c]{@{}l@{}}SCASeg\\ (MiT-B0)\end{tabular}}  & \CheckmarkBold & \XSolidBrush  & \XSolidBrush & \XSolidBrush  & 5.7 M  & 4.6 G  & 69.76  \\

& \CheckmarkBold & \CheckmarkBold  & \XSolidBrush & \XSolidBrush   & 5.7 M  & 4.9 G & 70.68 \color{green}(+0.92) \\
& \CheckmarkBold & \CheckmarkBold  & \CheckmarkBold & \XSolidBrush  & 5.7 M  & 5.2 G & 71.16 \color{green}(+1.40)\\
\rowcolor{gray!20}& \CheckmarkBold & \CheckmarkBold  & \CheckmarkBold & \CheckmarkBold   & 6.0 M  & 5.9 G & 72.35 \color{green}(+2.59) \\
\midrule
\multirow{4}{*}{\begin{tabular}[c]{@{}l@{}}SCASeg\\ (MSCAN-T)\end{tabular}} & \CheckmarkBold & \XSolidBrush  & \XSolidBrush & \XSolidBrush  & 6.3 M  & 6.0 G& 75.73 \\
& \CheckmarkBold & \CheckmarkBold  & \XSolidBrush & \XSolidBrush   & 6.3 M  & 6.3 G& 76.32 (\textcolor{green}{+0.59}) \\
& \CheckmarkBold & \CheckmarkBold & \CheckmarkBold & \XSolidBrush   & 6.3 M  & 6.6 G & 76.84 (\textcolor{green}{+1.11})\\
\rowcolor{gray!20}& \CheckmarkBold & \CheckmarkBold & \CheckmarkBold & \CheckmarkBold & 6.5 M  & 7.4 G & 77.88 (\textcolor{green}{+2.15})\\

\bottomrule
\end{tabular}
}
\label{crosslayer}
\end{center}
\end{table}

\subsection{Limitations and Future Work}
Despite the excellent performance and computational efficiency of our approach, there are some limitations related to parameter size, which may hinder its application in resource-constrained environments. In future work, we plan to explore techniques such as parameter pruning and knowledge distillation to develop a more compact and efficient version of our model, while maintaining its effectiveness. 

Our future work will focus on adaptive sampling of both tokens and channels to eliminate redundancy. Rather than compressing channels to a fixed number, we will enable the model to dynamically learn the effective channel count—per layer, per head, and even per sample—using rank-adaptive projections and sparsity-inducing gates (\textit{e.g.}, $L_0$/$L_1$ regularization or Gumbel–Softmax masking). For the tokens, we plan to use content-aware selection (differentiable Top-$k$ or budget-aware routing) to prioritize computation on salient regions while preserving dense gradients. 

\section{Conclusion}

This paper introduced Strip Cross-Attention (SCASeg), a novel decoder head tailored for semantic segmentation. We developed a Cross-Layer Block (CLB) that integrates hierarchical feature maps from various encoder and decoder stages to create a unified representation for Keys and Values. By incorporating the local perceptual strengths of convolution, the CLB enables SCASeg to effectively capture both global and local context dependencies across multiple layers, enhancing feature interaction at different scales and improving overall efficiency. Experimental results have shown SCASeg’s competitive performance across benchmark datasets.

Despite robust empirical results, edge deployment requires further efficiency improvements and reduced computational cost. While cross-layer connections are essential for hierarchical feature aggregation, they introduce significant inference overhead. To address this, we plan to (i) incorporate model-compression techniques, such as structured pruning and knowledge distillation, and (ii) implement self-learning mechanisms that adaptively select tokens and channels on demand, rather than using a fixed number, to develop a more robust and optimized segmentation model in the future.


\bibliographystyle{IEEEtran}
\bibliography{reference}

\vfill

\end{document}